\documentclass[sigconf, screen]{acmart}
\usepackage{longtable}
\usepackage{csquotes}
\MakeOuterQuote{"}
\AtBeginDocument{%
  \providecommand\BibTeX{{%
    \normalfont B\kern-0.5em{\scshape i\kern-0.25em b}\kern-0.8em\TeX}}}

\copyrightyear{2024}
\acmYear{2024}
\setcopyright{rightsretained}
\acmConference[C\&C '24]{Creativity and Cognition}{June 23--26, 2024}{Chicago, IL, USA}
\acmBooktitle{Creativity and Cognition (C\&C '24), June 23--26, 2024, Chicago, IL, USA}
\acmDOI{10.1145/3635636.3656186}
\acmISBN{979-8-4007-0485-7/24/06}

\usepackage{xcolor}
\colorlet{RED}{red}
\newcommand{\rev}[1] {{#1}}
\newcommand{\camera}[1] {{#1}}
\begin{document}

\title{Videogenic: Identifying Highlight Moments in Videos with Professional Photographs as a Prior}

% \author{Anonymous Author(s)}
\author{David Chuan-En Lin}
\affiliation{%
  \institution{Carnegie Mellon University}
  \streetaddress{5000 Forbes Ave.}
  \city{Pittsburgh, PA}
  \country{USA}
  }
\email{chuanenl@cs.cmu.edu}

\author{Fabian Caba Heilbron}
\affiliation{%
  \institution{Adobe Research}
  \streetaddress{}
  \city{San Jose, CA}
  \country{USA}
  }
\email{caba@adobe.com}

\author{Joon-Young Lee}
\affiliation{%
  \institution{Adobe Research}
  \streetaddress{}
  \city{San Jose, CA}
  \country{USA}
  }
\email{jolee@adobe.com}

\author{Oliver Wang}
\affiliation{%
  \institution{Adobe Research}
  \streetaddress{}
  \city{Seattle, WA}
  \country{USA}
  }
\email{owang@adobe.com}

\author{Nikolas Martelaro}
\affiliation{%
  \institution{Carnegie Mellon University}
  \streetaddress{5000 Forbes Ave.}
  \city{Pittsburgh, PA}
  \country{USA}
  }
\email{nikmart@cmu.edu}

\renewcommand{\shortauthors}{Lin et al.}

\begin{abstract}
  This paper investigates the challenge of extracting highlight moments from videos. To perform this task, we need to understand what constitutes a highlight for arbitrary video domains while at the same time being able to scale across different domains. Our key insight is that photographs taken by photographers tend to capture the most remarkable or \textit{photogenic} moments of an activity. Drawing on this insight, we present Videogenic, a technique capable of creating domain-specific highlight videos for a diverse range of domains. In a human evaluation study ($N$=50), we show that a high-quality photograph collection combined with CLIP-based retrieval (which uses a neural network with semantic knowledge of images) can serve as an excellent prior for finding video highlights. In a within-subjects expert study ($N$=12), we demonstrate the usefulness of Videogenic in helping video editors create highlight videos with lighter workload, shorter task completion time, and better usability.
\end{abstract}



% \textwidth
\rev{
\begin{teaserfigure}
  \centering
  \includegraphics[width=9cm]{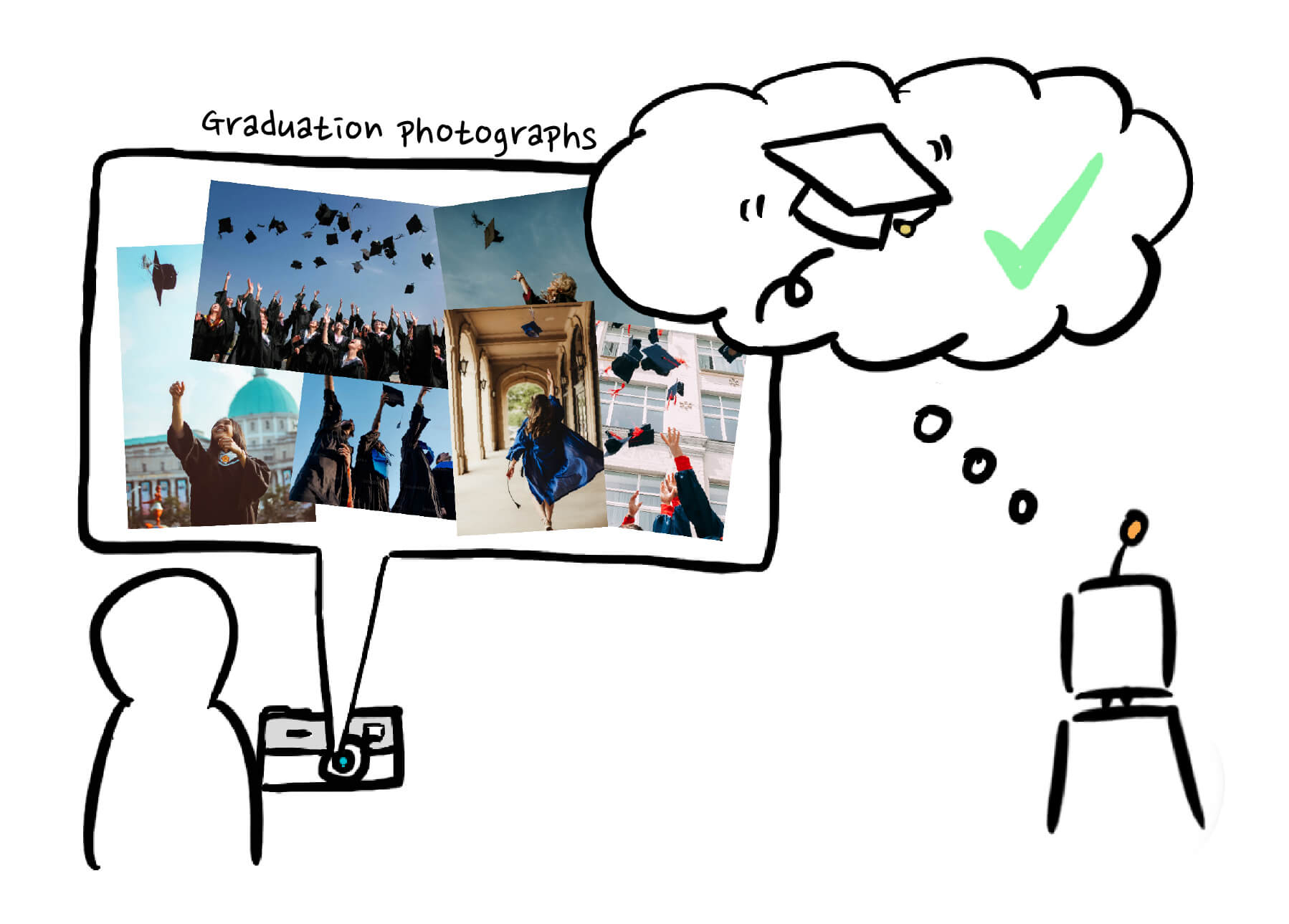}
  \caption{Videogenic leverages high-quality photographs as a prior to identify domain-specific highlights. For example, graduation photographs may frequently include people tossing their graduation caps.}
  \Description{Videogenic leverages high-quality photographs as a prior to identify domain-specific highlights. For example, graduation photographs may frequently include people tossing their graduation caps.}
  \label{fig:teaser}
\end{teaserfigure}
}

\maketitle
\section{Introduction}
Video highlight generation is the task of creating a short video clip that captures the highlight moments of a longer video or video collection. Such highlight videos can be useful for a variety of purposes. For example, people may wish to create short highlight clips of an activity (e.g., the moment of a great skateboard trick) or event (e.g., the main ceremony of a wedding) to share on social media. Video creators may wish to find good moments among large amounts of raw video footage to use for their videos. Video creators also may wish to upload short snippets of their longer videos on increasingly popular short-form video platforms such as TikTok, Instagram Reels, and YouTube Shorts to advertise their works to a larger audience \cite{short-video-app}. Video sharing platforms may wish to let users see short previews of videos before watching them (e.g., YouTube plays a 3-second preview when the user hovers over a video's thumbnail \cite{youtube-preview}).

Many video highlight generation approaches have been proposed to support the demand for highlight videos. Since the definition of what constitutes a "highlight" is highly dependent on the domain of interest (e.g., a skateboard trick or a cool dance move), a key challenge of video highlight generation is to find some way of encoding domain knowledge about what a good highlight is within the system. 
Several works make use of domain-specific features to identify highlights, such as detecting the presence of people \cite{lee2012discovering} or identifying when a goal is scored in sports videos \cite{yow1995analysis}. However, such systems only work for the specific domain that they are designed for. More recent works make use of neural networks to learn a model of video highlights from data, such as from pairs of highlight videos and their source videos \cite{sun2014ranking} or videos with particular segments labeled as highlights \cite{yao2016highlight}. Nonetheless, such systems require resource-intensive training of models on large amounts of data.

\vspace{0.3cm}
\begin{quotation}
\textit{``Photography is the simultaneous recognition, in a fraction of a second, of the significance of an event.''}
\vspace{0.1cm}
\par\raggedleft--- \textup{Henri Cartier-Bresson, Photographer}
\end{quotation}
\vspace{0.3cm}

In this research, we take a different approach to creating highlight clips from longer videos by leveraging the \textit{domain knowledge of photographers}. 
We posit that pictures of events or activities taken by photographers capture the most remarkable or \textit{photogenic} moments of an activity. Given an arbitrary domain of interest, we search for a small collection of professional photographs depicting the activity and create an average representation of these photographs. We then compare the average representation against each frame of the source video to compute similarities.
High similarity between the video frame and the average photograph representation corresponds to a high "highlight score".
We show that a high-quality photograph collection combined with the use of a semantic model to encode its representation (CLIP \cite{clip}) can serve as a strong prior for finding video highlights in arbitrary domains without any further training. In addition, our method also \rev{\textit{implicitly encodes useful photography knowledge}}, such as good composition and framing.

In this paper, we present Videogenic, a \rev{technique for identifying} highlight \rev{moments in} videos by leveraging a small sample of photographs that represent an arbitrary activity or event. We first introduce a set of principles to ground the \rev{task of} domain-agnostic highlight generation. \rev{We build an interactive proof-of-concept system for creating highlight videos to validate our technique}. We test the \rev{performance} of \rev{our technique} against a baseline by generating highlight videos for a variety of source videos covering various domains and asking people to pick their preferences. In an expert study, we further evaluate the usefulness of Videogenic for video editors against a baseline, demonstrating improvements in workload, usability, and task completion time. \rev{We asked external raters to evaluate the videos created by the editors, demonstrating strong "highlightness" and production quality for the highlight videos created with Videogenic.}
% \vfill\break

This research thus makes the following contributions:
\begin{itemize}
\item{\rev{\textbf{Videogenic, a simple yet effective technique for identifying highlight moments in videos.} By leveraging professional photographs as a prior and leveraging CLIP's representation power, our technique scales to \textit{arbitrary} video domains \textit{out-of-the-box}. We built a proof-of-concept system to validate our technique with professional video creators.}}
\item{A \textbf{human evaluation} ($N$=50 participants) of the \rev{performance} of Videogenic against a baseline. Participants preferred Videogenic's highlights on average 80\% of the time.}
\item{An \textbf{expert study} ($N$=12 video editors) evaluating the usefulness of Videogenic against a manual editing baseline.} Participants experienced lighter workload, shorter task completion time, and better usability when using Videogenic. External raters rated strong "highlightness" and production quality for the highlight videos created with Videogenic.
\end{itemize}

\section{Related Work}

\rev{Our work is situated among extensive literature on video highlight generation. We discuss works that adopt heuristics-based approaches, data-driven approaches, as well as HCI approaches.}

\subsection{Heuristic Approaches}
As the content within videos can vary considerably across different domains, many works focus on a single video domain to define a set of domain-specific heuristics.
A large body of work focuses on sports videos. A sports game usually has a well-defined structure and consists of various stages. Among the various stages, only a small selection contains highlight moments. For example, the highlights of a soccer video are generally the stages in which the goals are scored. Since the relevant highlight stages can also vary across different types of sports, a variety of approaches have been proposed across various sports categories, such as for soccer \cite{yow1995analysis}, baseball \cite{rui2000automatically}, basketball \cite{nepal2001automatic}, and cricket \cite{kolekar2006event}. In addition, a growing body of work addresses video highlight generation for egocentric videos, possibly due to a rise in popularity of personal action cameras \cite{insta360, gopro}. These works often make use of various pre-defined cues, such as the detection of people, faces, and objects  \cite{lee2012discovering, lu2013story}. Various commercial software also rely on pre-defined heuristics. For example, Insta360's FlashCut feature \cite{flashcut} uses image recognition models to detect hands and faces.
\rev{Song et al. \cite{song2016click} uses a variety of aesthetic heuristics to select thumbnails for videos.}
In this research, we avoid using pre-defined heuristics and make use of the latent knowledge encoded within professional photographs.

\subsection{Data-Driven Approaches}
Following recent advancements in machine learning research, recent works investigate data-driven methods for video highlight generation. For example, Yao et al. \cite{yao2016highlight} learn from a dataset consisting of long videos segmented into various highlight and non-highlight segments. Nonetheless, video datasets with labeled highlights are difficult to collect as highlights are subjective and require human annotation. Moreover, the annotation task can be ambiguous since what constitutes a highlight is dependent on the video domain of interest. Looking at sports videos as an example, the human annotator would need to understand the rules of the particular sport to be able to make the annotations. As a result, the annotated datasets may contain noisy labels. Thus, several researchers have looked into exploiting \textit{proxy} priors. Examples include using pairs of edited videos and their raw videos \cite{sun2014ranking}, pairs of GIFs and their video sources \cite{gygli2016video2gif}, short user-generated videos \cite{xiong2019less}, web-images \cite{khosla2013large, kim2018exploiting}, \rev{titles \cite{song2015tvsum},} and detecting shared visual events across multiple videos \cite{chu2015video}. Our work is closely related to this thread of work. However, prior works involve expensive model training on (noisy) large-scale \rev{video or image} data. In contrast, we show how only a small set of high-quality photographs from photographers combined with the semantic encodings of CLIP \cite{clip} can act as an excellent prior for creating highlight videos. This allows our technique to work \textit{out-of-the-box} on arbitrary videos with no additional training necessary.

\subsection{\rev{HCI Approaches}}
\rev{The HCI community has explored methods of obtaining domain knowledge information from people to support finding highlights in videos. Several past works adopt a crowdsourcing approach. For example, Wu et al. \cite{wu2011video} leverage crowd wisdom for summarizing videos, capable of adapting to various video domains and summary abstraction levels. Bernstein et al. \cite{bernstein2011crowds} use synchronous crowds to crowdsource highlight moments within videos in real-time. Instead of recruiting crowd workers, several works leverage existing users within social communities of video content. San et al. \cite{san2009you} identify recurring scenes within a video domain on video-sharing sites as a form of "social summarization." Sun et al. \cite{sun2016videoforest} convert video and user comments into tree-like visual summaries to help people identify content highlights. Yang et al. \cite{yang2022catchlive} provide real-time summaries of livestreams based on streaming content and user interaction data. In this research, we follow a similar spirit, by tapping into the domain knowledge of professional photographers through their photography as a way of finding highlights.} 

\rev{HCI researchers have also built interfaces to better support browsing highlight moments in videos. Matejka et al. \cite{matejka2013swifter} use a grid of thumbnails designed to support users with easier scrubbing and selection of video moments. Matejka et al. \cite{matejka2014video} also help users quickly find relevant video sections through large collections of videos using text-based timelines and associated metadata about what is in the video. In this research, we support users by scrubbing through an interface with predicted highlight score data and allow people to use a brush feature across video segments to create their final highlight video.}

\section{Principles of Video Highlight Generation}
We define three key design principles for the task of generating video highlights based on prior work to ground the development of Videogenic.

\subsection{Principle 1: Domain-Agnostic Highlights}
Our first principle is to create a system that is able to scale to a diverse range of domains. To achieve this, our system should be established on a \textit{domain-agnostic prior} (i.e., does not limit the system to the specific domain it was designed for). In this research, we use the domain-agnostic prior of professional photographs and demonstrate flexibility across sports videos, events videos, nature videos, and more.

\subsection{Principle 2: Domain-Specific Knowledge}
Our second principle is to design the system such that is able to understand what constitutes a highlight for the specific domain of interest \cite{sun2014ranking}. For example, the highlights of a skateboarding video (e.g., the impressive skateboard trick) can be very different from the highlights of a wedding video (e.g., when the officiant addresses the marriage partners). We show how photographs taken by photographers can encode rich domain-specific knowledge in addition to encouraging good composition and framing.

\subsection{Principle 3: Flexible User Control}
Our third principle is to give users flexible controllability. For example, users may wish to interpret the highlight predictions \cite{krause2016interacting}, correct errors \cite{amershi2019guidelines}, select their preference among multiple feasible highlights, or add individual touches \cite{pseudoclient}. To support this, we should give users multiple modes to pick from at various stages of the system. By selecting the automatic mode in all stages, the user may create highlight videos in a completely automatic fashion. Alternatively, the user may opt for an interactive mode to have greater fine-grained control over the final output.

\section{Implementation}
\label{section:implementation}
Our three principles are manifested in Videogenic and guide its implementation. In this section, we detail the implementation of our system, including (1) selecting the topic, (2) computing highlight scores, and (3) generating the highlight video.

\subsection{Classifying the Activity}
Our first step is determining the primary activity of the video. We offer two methods for the user: (1) by providing a keyword and (2) by using an automatic classifier.

\begin{figure*}[tbp]
  \centering
  \includegraphics[width=14cm]{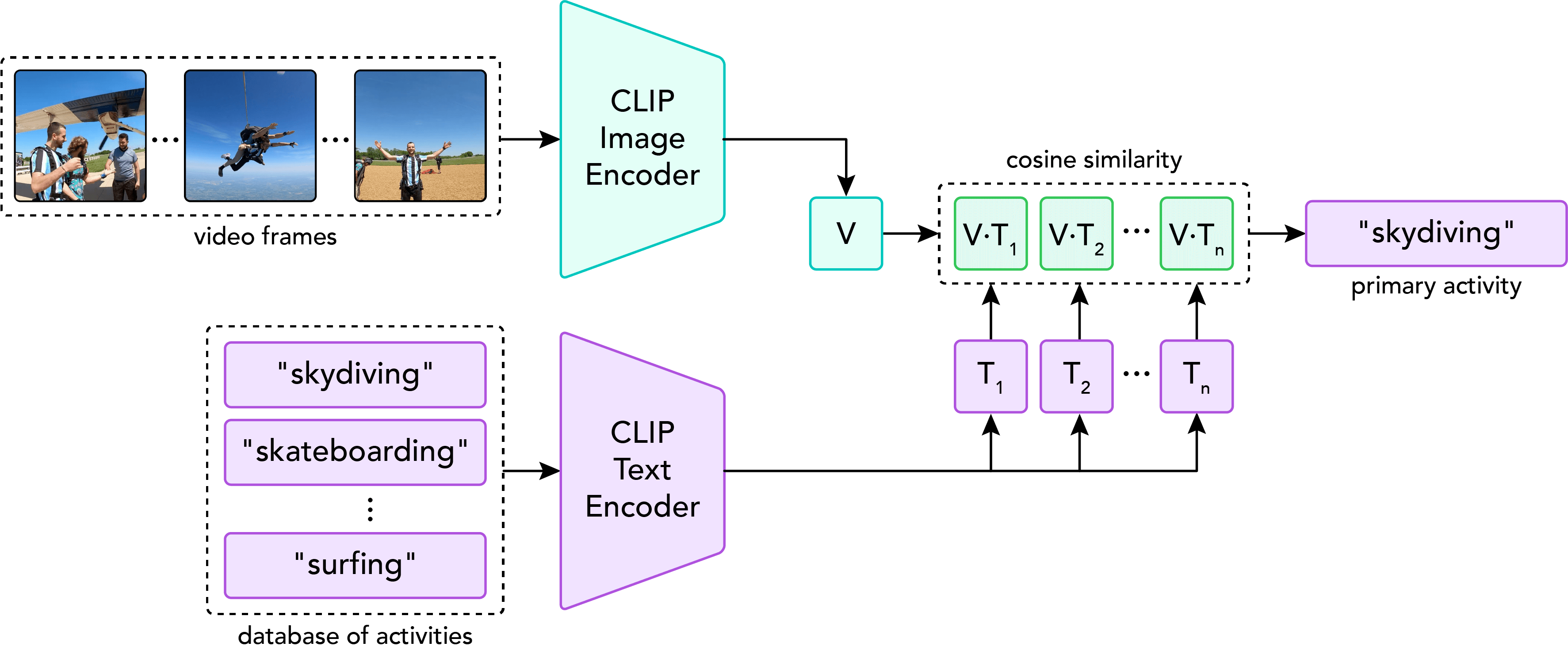}
  \caption{Automatic classifier. Given the frames of a video and a database of activity labels, Videogenic performs pairwise comparisons to predict the primary activity of the video.}
  \Description{Automatic classifier. Given the frames of a video and a database of activity labels, Videogenic performs pairwise comparisons to predict the primary activity of the video.}
  \label{fig:classify}
\end{figure*}

\begin{figure*}[tbp]
  \centering
  \includegraphics[width=13cm]{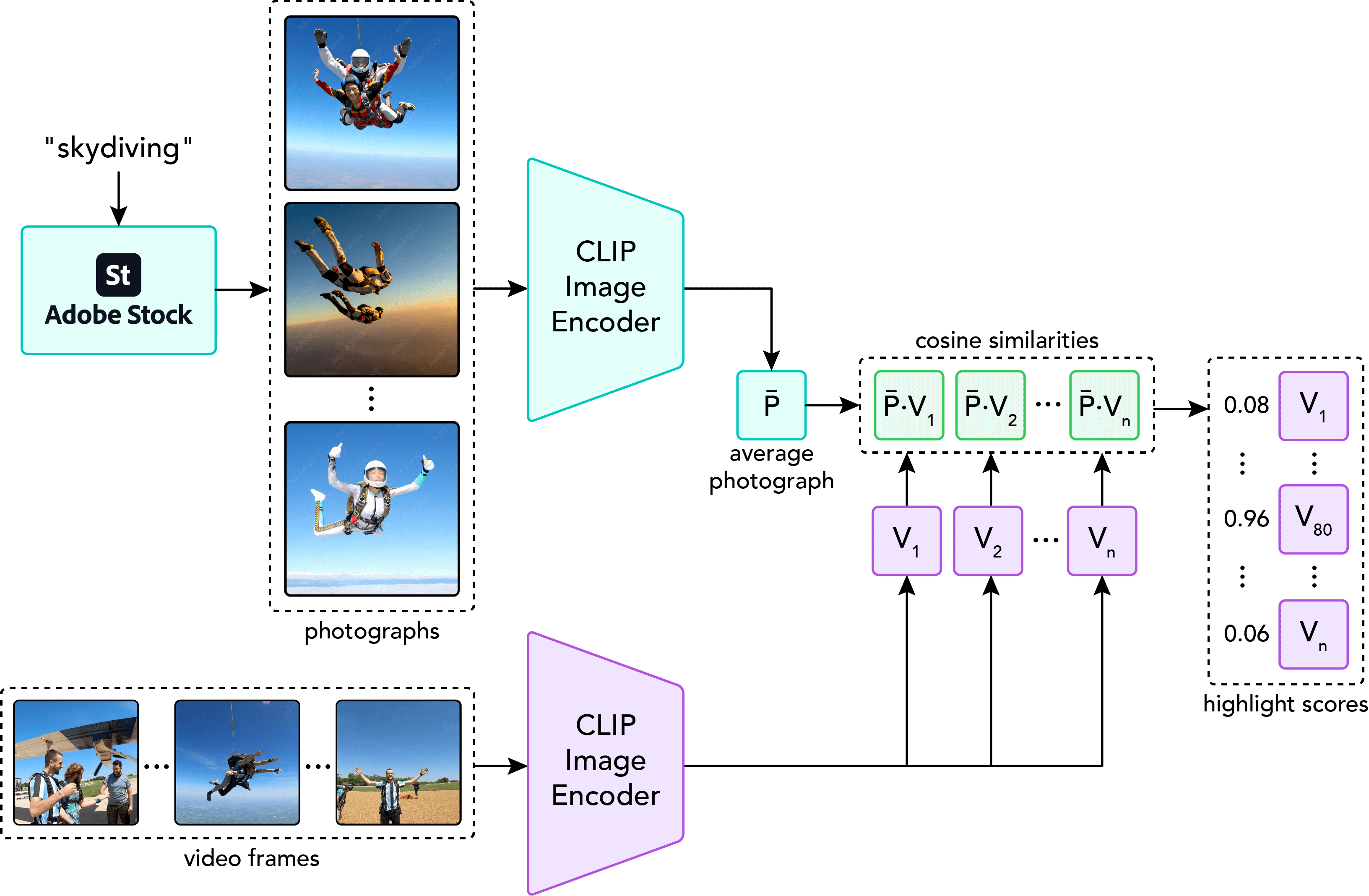}
  \caption{Computing highlight scores. Given an activity label (e.g., \texttt{skydiving}), Videogenic retrieves 10 stock photographs and computes the average photograph representation. Given each frame of a video and the average photograph, Videogenic performs pairwise comparisons to predict a highlight score for each frame.}
  \Description{Computing highlight scores. Given an activity label (e.g., skydiving), Videogenic retrieves 10 photographs from Adobe Stock and computes its average photograph representation. Given each frame of a video and the average photograph, Videogenic performs pairwise comparisons to predict a highlight score for each frame.}
  \label{fig:compute}
\end{figure*}

\subsubsection{User-Specified Keyword} 
\begin{figure*}[tbp]
  \centering
  \includegraphics[width=12cm]{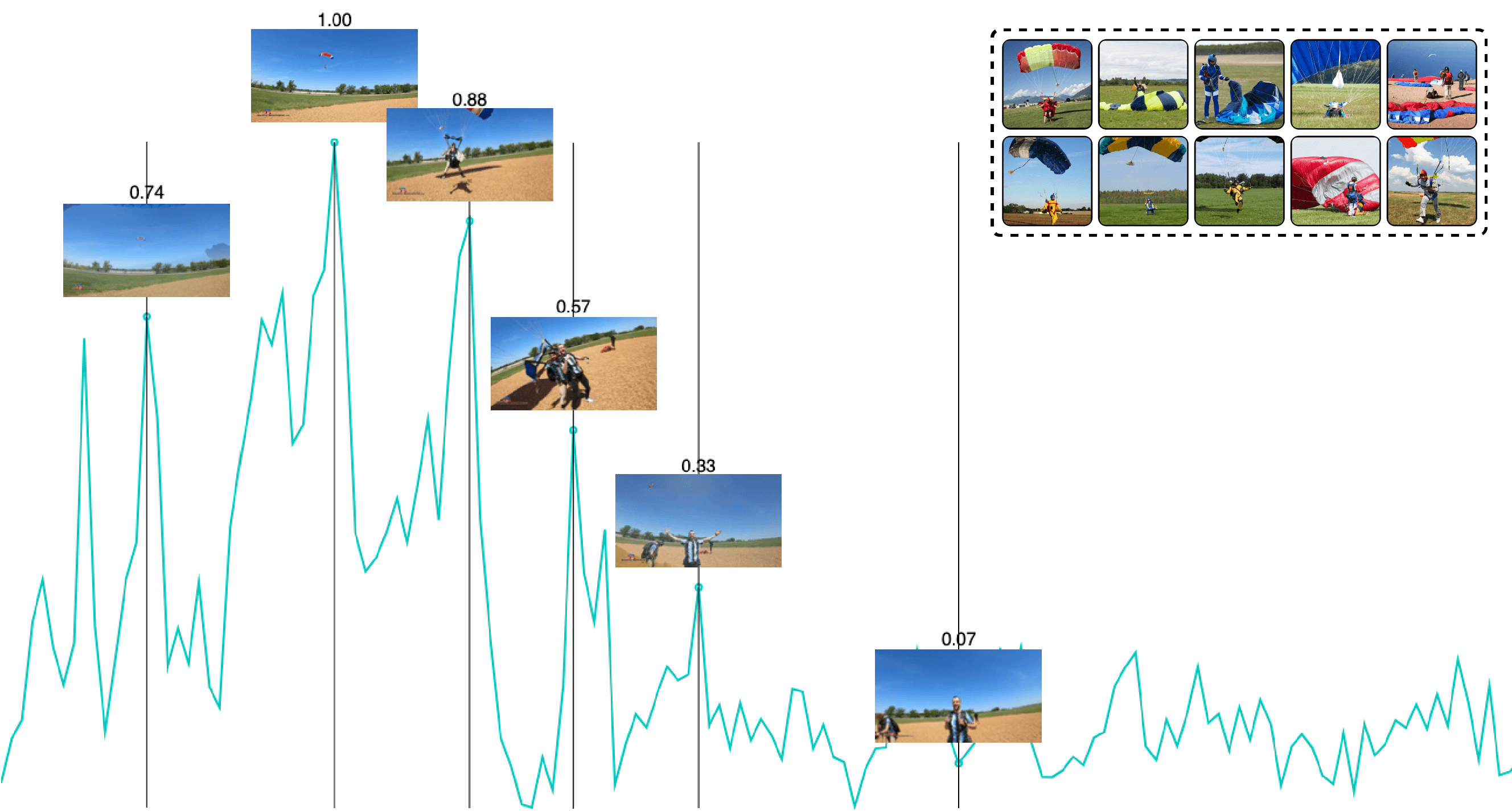}
  \caption{Example video frames and their corresponding highlight scores within a long skydiving video, using the keyword \texttt{skydiving landing}. The top-right corner displays the photograph collection used by Videogenic.}
  \Description{Example video frames and their corresponding highlight scores within a long skydiving video, using the keyword "skydiving landing". The top-right corner displays the photograph collection used by Videogenic.}
  \label{fig:skydiving-landing-highlights}
\end{figure*}

The user may specify a keyword (e.g., \verb|skydiving|) as the primary activity. This gives the user the flexibility to experiment with different keywords (Principle 3).
and customize the highlight video (Principle 3).
For example, given video footage of skydiving, the user may give the keyword of \verb|skydiving landing| to create a highlight video for skydiving touchdowns (Figure \ref{fig:skydiving-landing-highlights}).

\subsubsection{Automatic Classifier}
To support an automatic mode, the user may allow our system to automatically determine the primary activity of the video (Figure \ref{fig:classify}). Given a video, we encode each video frame through the CLIP \cite{clip} image encoder to produce its semantic representation. We then concatenate the frame-wise representations into the representation for the video. Given a database of various activity categories (e.g., skydiving, skateboarding, surfing), we encode each activity label through the CLIP text encoder. We then compare each encoded activity label representation with the encoded video representation via cosine similarities, and select the top-ranking activity label as the primary activity.

\begin{figure*}[tbp]
  \centering
  \includegraphics[width=14cm]{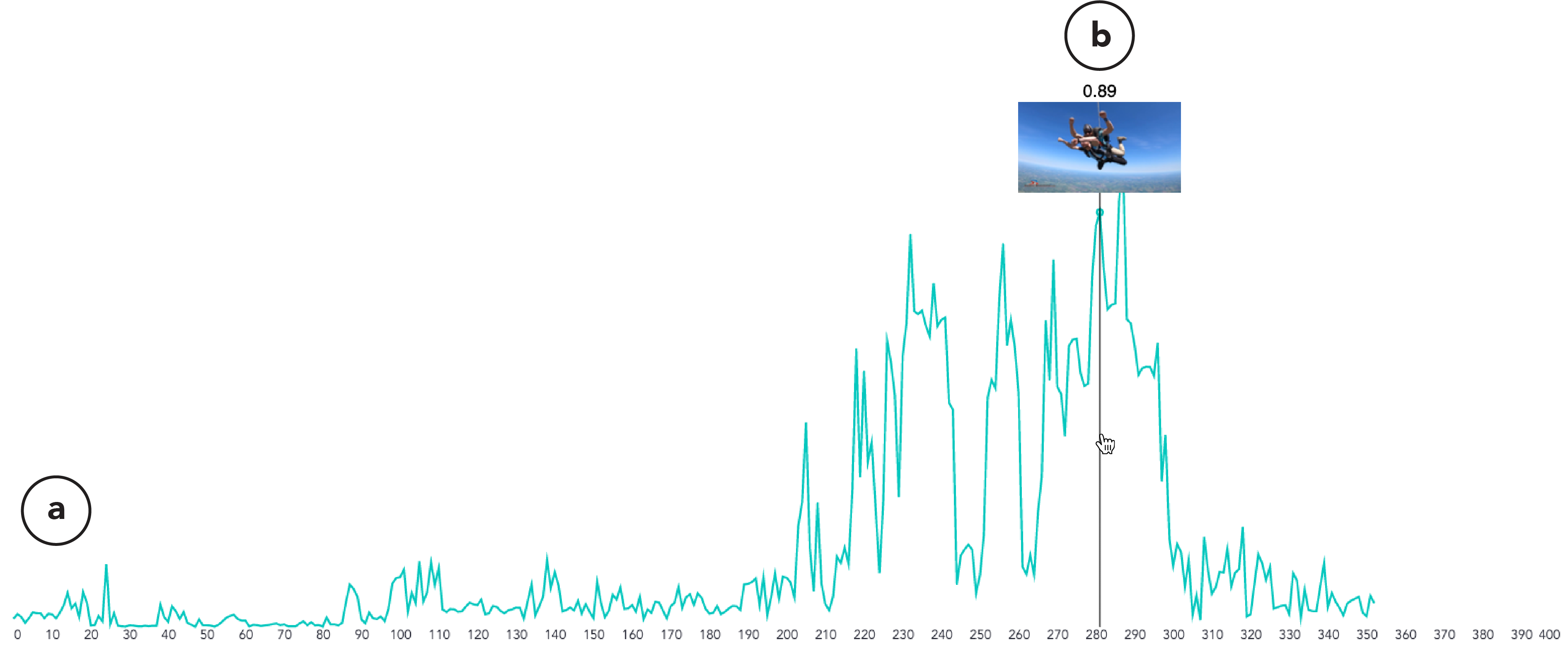}
  \caption{Highlight graph. The highlight graph visualizes the distribution of predicted highlight scores across the video (a). The user may scrub through the graph to inspect a corresponding video frame and its highlight score (b).}
  \Description{Highlight graph. The highlight graph visualizes the distribution of predicted highlight scores across the video (a). The user may scrub through the graph to inspect a corresponding video frame and its highlight score (b).}
  \label{fig:highlight-graph}
\end{figure*}

\begin{figure*}[tbp]
  \centering
  \includegraphics[width=14cm]{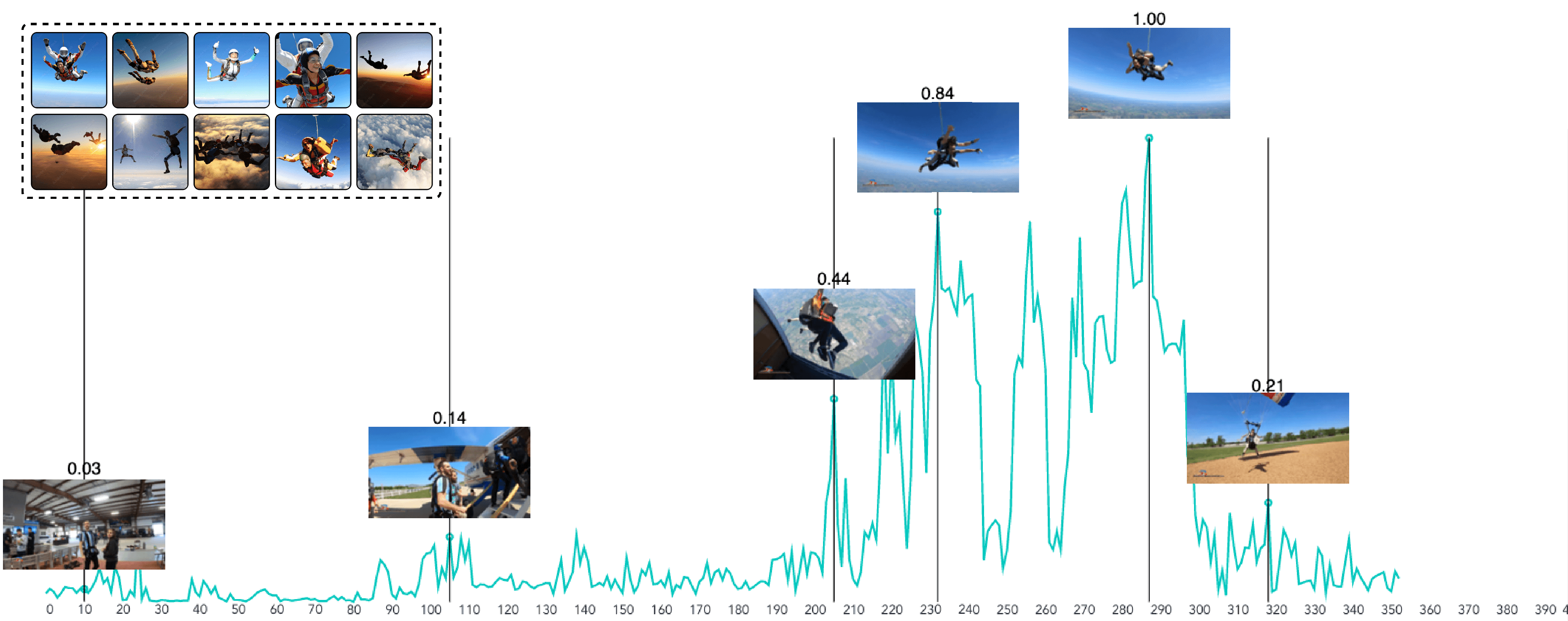}
  \caption{Example video frames and their corresponding highlight scores within a long skydiving video, using the keyword \texttt{skydiving}. The top-left corner displays the photograph collection used by Videogenic.}
  \Description{Example video frames and their corresponding highlight scores within a long skydiving video, using the keyword "skydiving". The top-right corner displays the photograph collection used by Videogenic.}
  \label{fig:skydiving-highlights}
\end{figure*}

\begin{figure*}[tbp]
  \centering
  \includegraphics[width=13cm]{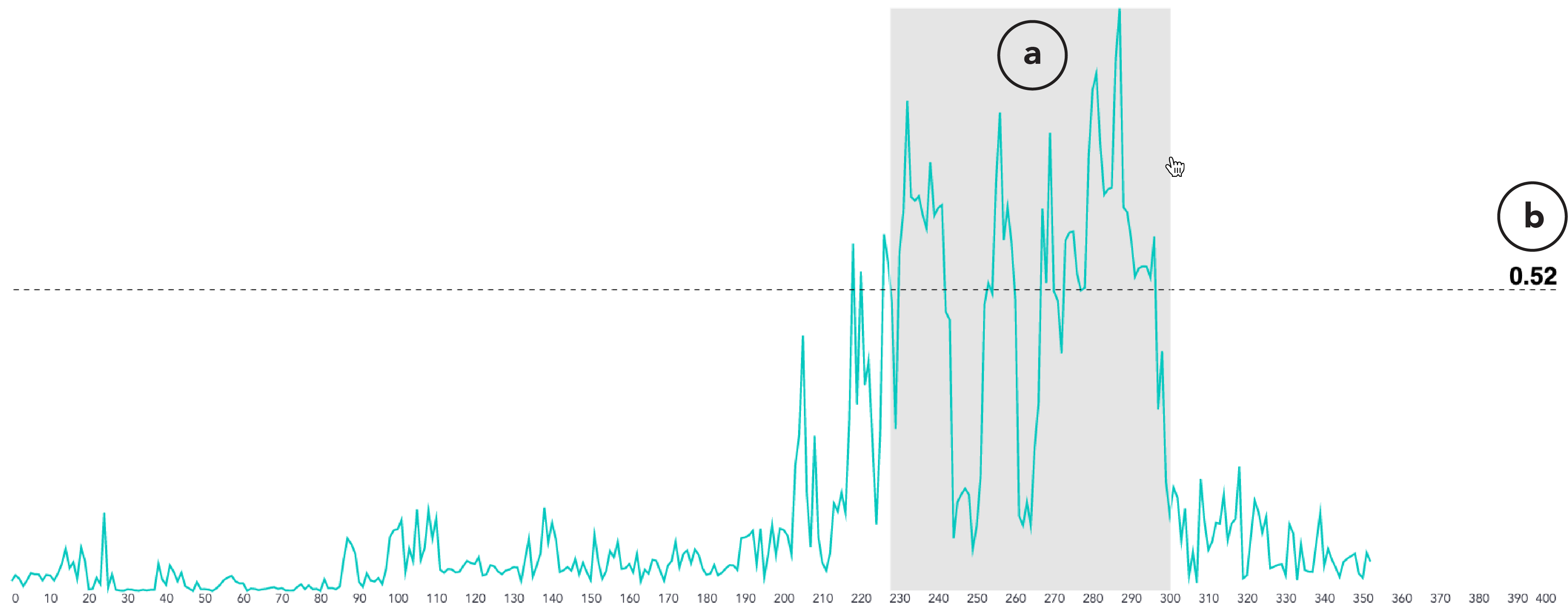}
  \caption{The user may brush through the highlight graph to select an interval of the video to use for the highlight video (a). The interface displays a dashed line and a text label to indicate the average highlight value of the selected interval (b).}
  \Description{The user may brush through the highlight graph to select an interval of the video to use for the highlight video (a). The interface displays a dashed line and a text label to indicate the average highlight value of the selected interval (b).}
  \label{fig:highlight-brush}
\end{figure*}

\subsection{Computing Highlight Scores}
The second step is computing frame-wise highlight scores for the video (Figure \ref{fig:compute}). Given the primary activity (e.g., \verb|skydiving|), Videogenic automatically retrieves 10 professional photographs depicting the activity from a database of professional photography \camera{(e.g., top 10 photos from Adobe Stock search results)}. We encode each photograph through the CLIP image encoder ($\mathbf{P}$). We then average all the photographs' representations $\mathbf{P_{1,...,10}}$ to create a representation for the average photograph ($\bar{\mathbf{P}}$), which we use as the prior for judging the highlight scores of each video frame. Our intuition is that professional photographs capture the most highlight-worthy moments of an activity with skillful composition and framing. By exploiting the domain knowledge of photographers (Principle 2), we are not using any domain-specific priors (e.g., detecting people or faces), making our system scalable across diverse domains (Principle 1) through the use of new sets of photographs to compute the average photograph for different domains (e.g., nature photographs or wedding photographs).
\rev{We empirically experimented with using different numbers of photographs. We found that using a single photograph sometimes introduces irrelevant attributes to the highlight. For example, the photograph is taken during a particular time of day, includes a particular background, or depicts a subject of a particular gender. We found that creating an \textit{average photograph} from ten photographs successfully removes the effects of irrelevant attributes.} Next, we encode the video frames through the CLIP image encoder ($\mathbf{V}$). We then compare distance the average photograph representation with each encoded video frame via cosine similarities (Eq. \ref{eqn:eq1}). This gives us a vector of highlight scores ($\mathbf{H}$) for each video frame. Finally, we normalize the highlight scores across the video on a scale of $[0, 1]$.

\begin{equation}
\label{eqn:eq1}
    \mathbf{H} = \bar{\mathbf{P}} \cdot \mathbf{V}^\intercal
\end{equation}

\subsubsection{Highlight Graph}
We provide users with an interface to visualize the distribution of highlight scores across the video (Principle 3) (Figure \ref{fig:highlight-graph}). This allows users to easily identify potential highlights in the video at a glance.
We plot the highlight scores of each frame, where the y-axis represents the normalized highlight score and the x-axis represents the ID of each video frame in chronological order (Figure \ref{fig:highlight-graph}a).
Analogous to the playhead of a video player, the user may scrub through the visualization to inspect the corresponding video frame thumbnail and highlight score (Figure \ref{fig:highlight-graph}b). In Figure \ref{fig:skydiving-highlights}, we show several example moments in a skydiving video and its highlight scores (with \verb|skydiving| as the primary activity). We see that the moments of freefall have the highest scores, the moments of jumping out of the plane and landing have moderate scores, and the moments of preparation and boarding the plane have low scores. By changing the keyword to \verb|skydiving landing|, we show several example highlight moments based on a new set of images that were sampled to create the average image (Figure \ref{fig:skydiving-landing-highlights}). Allowing simple changes of the keyword can allow users to explore different kinds of highlights within the same video.

% \begin{figure*}[tbp]
%   \centering
%   \includegraphics[width=13cm]{figures/videogenic-generate.png}
%   \caption{Highlight template pipeline. Given the highlight scores for each frame of the video and a desired highlight video length $N$ (e.g., 10 seconds), Videogenic finds an $N$-length interval within the video that has the maximum sum of highlight scores. Videogenic then adds video effects and music as defined by a selected highlight template to produce the final highlight video.}
%   \Description{Highlight template pipeline. Given the highlight scores for each frame of the video and a desired highlight video length N (e.g., 10 seconds), Videogenic finds an N-length interval within the video that has the maximum sum of highlight scores. Videogenic then adds video effects and music as defined by a selected highlight template to produce the final highlight video.}
%   \label{fig:generate}
% \end{figure*}

\subsection{Generating the Highlight Video}
Our final step is generating the highlight video. We offer two methods for the user: (1) by a user-selected interval in the highlight visualization and (2) by automatically identifying an interval with high scores.

\label{section:user-selection}
\subsubsection{User Selection} The user may select an interval of the source video as their highlight video by brushing through the highlight visualization (Figure \ref{fig:highlight-brush}a). As the user brushes through the visualization, we display the average highlight score of the selected interval (Figure \ref{fig:highlight-brush}b). We then output the user-selected interval as the final highlight video.

\subsubsection{Automatic Selection}

% \rev{We provide a set of highlight templates for users to automatically create highlight videos. The highlight templates are created by one of the authors based on several popular TikTok \cite{tiktok-growth} trends and include pre-made video effects and music. Note that the templates themselves are not the primary contribution of this paper and serve mainly as a possible workflow for highlight video creations common on social media platforms. We create several templates for users to choose from, including various music genres (e.g., upbeat, hype, tranquil) and lengths (e.g., 7 to 18 seconds)}. After the user selects a template,
We search for a continuous interval of length $N$ (e.g., 10 seconds) within the video for an interval that has the maximum sum of highlight scores by sliding a window across the frame-wise highlight scores to find the maximum subarray. 

\begin{figure*}[tbp]
  \centering
  \includegraphics[width=12cm]{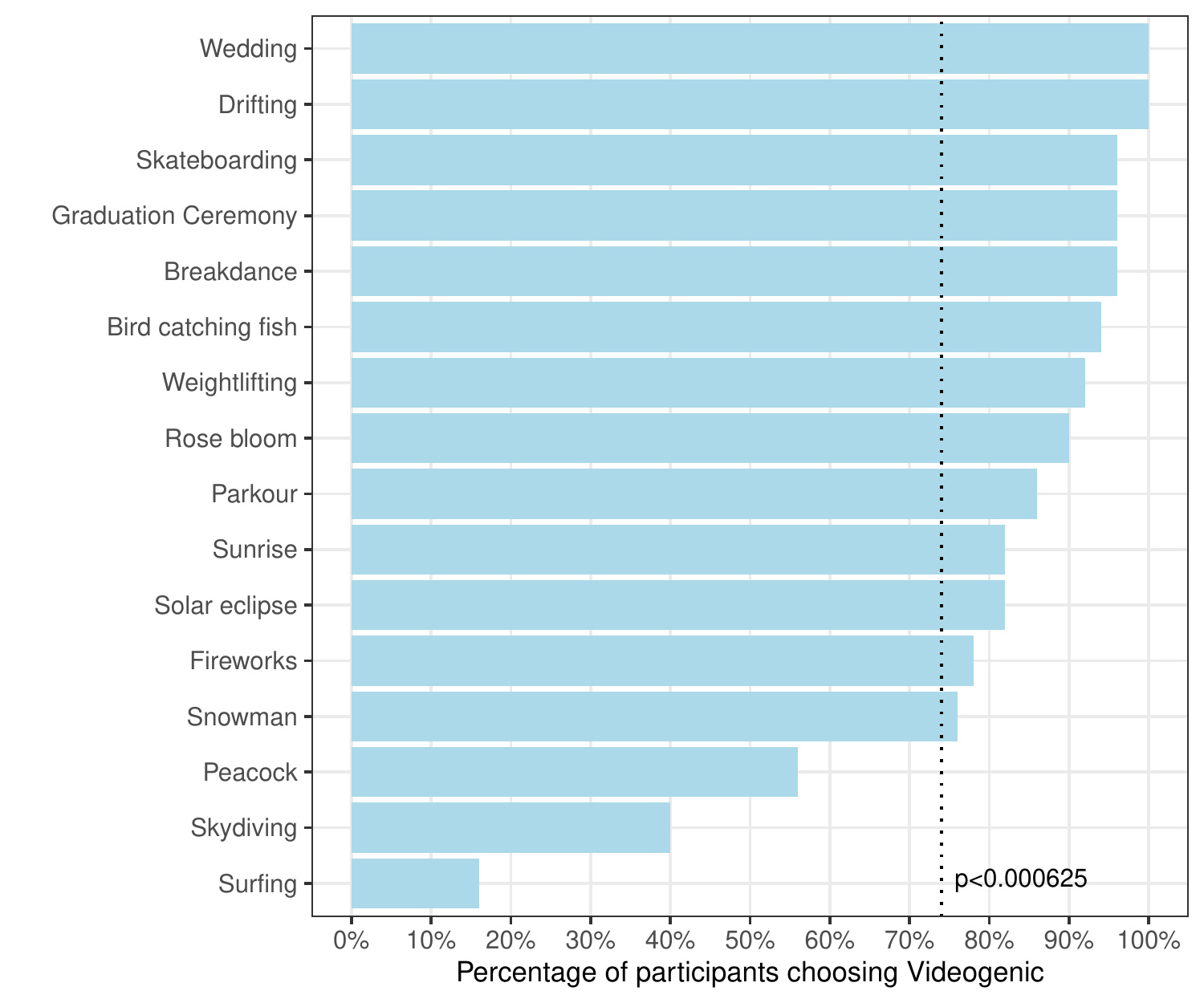}
  \caption{Human evaluation results ($N=50$). The y-axis lists the videos in the evaluation study. The x-axis shows the percentage of participants who preferred Videogenic's highlight over the baseline's. The dashed line marks the point of statistical significance ($p$<0.000625). The baseline is based on CLIP \cite{clip} text-video similarity.}
  \Description{Human evaluation results (N=50). The y-axis lists the videos in the evaluation study. The x-axis shows the percentage of participants who preferred Videogenic's highlight over the baseline's. The dashed line marks the point of statistical significance (p<0.000625). The baseline is based on CLIP text-video similarity.}
  \label{fig:human-evaluation-results}
\end{figure*}

\begin{figure*}[tbp]
  \centering
  \includegraphics[width=15cm]{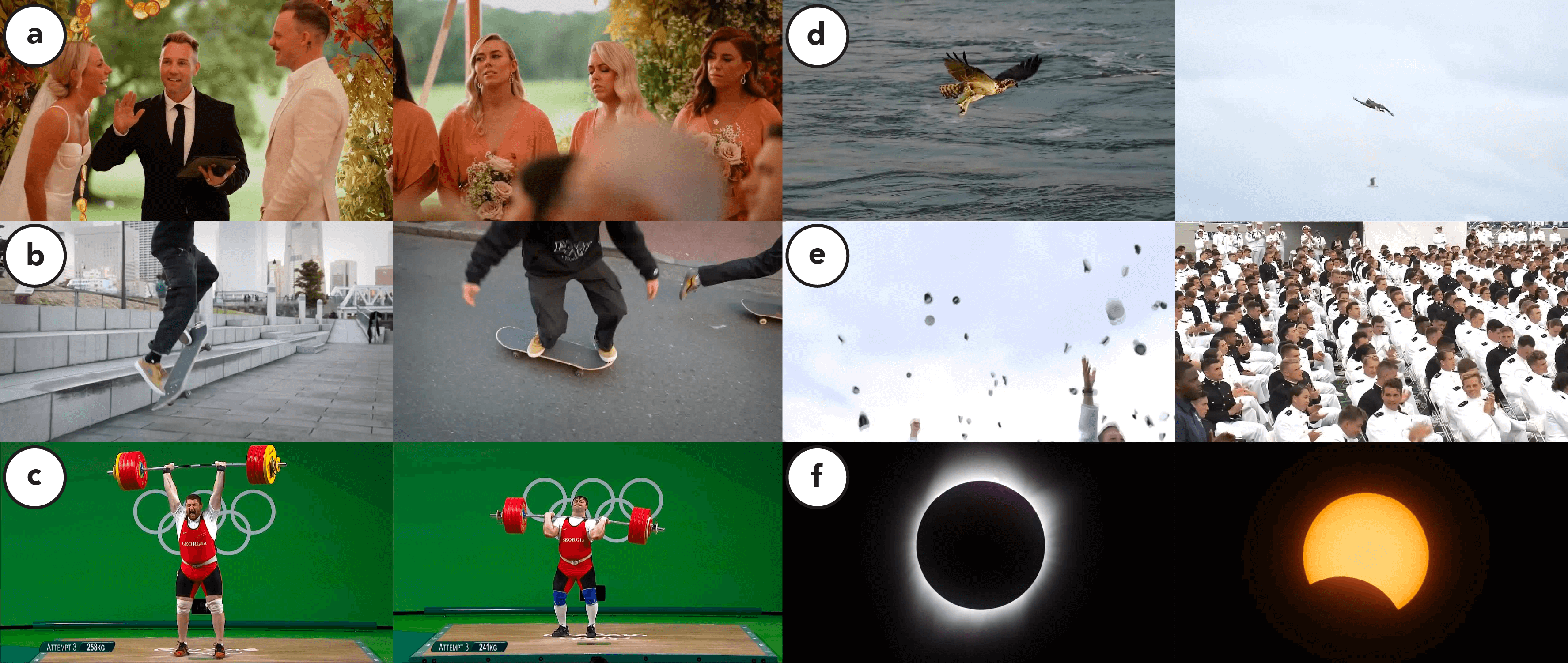}
  \caption{Example qualitative human evaluation results for wedding (a), skateboarding (b), weightlifting (c), bird hunting fish (d), graduation ceremony (e), and solar eclipse (f). For each pair, the left shows the highlight by Videogenic and the right shows the highlight by the baseline. Videogenic identifies the most remarkable moments with good composition and framing.}
  \Description{Example qualitative human evaluation results for wedding (a), skateboarding (b), weightlifting (c), bird hunting fish (d), graduation ceremony (e), and solar eclipse (f). For each pair, the left shows the highlight by Videogenic and the right shows the highlight by the baseline. Videogenic identifies the most remarkable moments with good composition and framing.}
  \label{fig:human-evaluation-examples}
\end{figure*}

\section{Human Evaluation}
\label{section:human-evaluation}
To test the functionality of Videogenic, we run a human evaluation study on video highlights generated with Videogenic and a strong baseline method of CLIP similarity between text and video frames. The following outlines our experimental setup, procedure, and results.

\subsection{Setup}

\subsubsection{Source Video Collection}
We first collect our set of source videos that we would like to generate highlight videos for. To test the flexibility of Videogenic, we collect 16 diverse source videos from YouTube of (1) various \textit{lengths} ranging from 30 seconds to 4 hours (mean=1,823 seconds, SD=3,584 seconds), (2) various \textit{formats} such as live broadcasts (e.g., drift racing live broadcast), timelapses (e.g., sunrise timelapse), vlogs (e.g., a day in the life of a surfer vlog), documentaries (e.g., peacock documentary), and unedited videos (e.g., unedited fireworks show) and (3) various \textit{categories} such as sports (e.g., skateboarding, weightlifting, parkour), events (e.g., wedding, graduation ceremony, building snowman), nature (e.g., sunrise, solar eclipse, rose bloom), and animals (e.g., bird hunting fish, peacock courtship).

\subsubsection{Videogenic and Baseline Setup}
For each collected video, we generate a highlight video for it with two systems: Videogenic and a baseline system.

\textbf{Videogenic Setup.}
We automatically generate a highlight video for each collected video with Videogenic as detailed in Section \ref{section:implementation}.

\textbf{Baseline Setup.}
We adopt CLIP similarity between \textit{text} and video frames \cite{which-frame} as our baseline system. Specifically, rather than computing an average photograph representation to compare against video frames, the method compares the keyword against video frames directly via CLIP cosine similarities. Given a keyword, the method is to find semantically relevant video frames with strong performance. We make use of this method as our baseline for two reasons. First, like Videogenic, the method is able to perform in a zero-shot manner with no training required. This matches our key principle of creating a system that can be used \textit{out-of-the-box} to generate video highlights for arbitrary videos (Principle 1). Second, a comparison between Videogenic and the method can help examine the usefulness of using \textit{professional photographs} as our prior. To create our highlight video, we determine the primary activity keyword as shown in Figure \ref{fig:classify}. We then use the baseline method to compute scores for each video frame based on the keyword and take an interval with the maximum sum of scores as the highlight video.

\subsection{Procedure}
We run a human evaluation study on Prolific \cite{prolific} to evaluate the highlight videos generated by Videogenic and the baseline system.
We recruit 50 US-based participants with standard sampling and prescreen participants such that they must have experience in using TikTok, a short-form video platform, to ensure that participants are familiar with the concept of highlight videos.
After receiving participants' consent, we ask each participant to compare the highlight videos generated with the two conditions for each of the 16 sample videos in a paired comparison, two-alternative forced choice manner \cite{2afc}.
% The participant does not know which highlight video is generated with Videogenic or the baseline system and we randomize the left and right orderings of how the videos are displayed.
% To ensure the quality of the responses, we have a check to make sure participants are able to see the videos correctly and we include two quality control questions \cite{agley2022quality}.
The study takes approximately 3 minutes to complete and we compensate participants \$2 USD for their time.
% , which is above the US minimum wage.

\subsection{Results}

Overall, participants prefer the highlights by Videogenic over the baseline system for 14 out of 16 videos (preference determined by the majority) (Figure \ref{fig:human-evaluation-results}). We further analyze the results for statistical significance through a binomial test with Bonferroni correction (16 tests, significance level at $\alpha$<$\frac{0.01}{16}$=0.000625). Participants significantly prefer the highlight videos generated with Videogenic (mean=80.00\%, SD=9.45\%, $p$<0.000625) compared to the baseline.

Figure \ref{fig:human-evaluation-examples} shows qualitative examples of highlights identified by the Videogenic (left) versus the baseline (right). We see that, in general, the baseline method is able to correctly identify relevant content within diverse videos. For example, the skateboarding source video includes various irrelevant content such as the skateboarders eating, shopping, playing arcade games, and walking around the city. However, we see that Videogenic is able to more accurately identify the most remarkable highlight moments, given professional photographs as a prior. In the figure, Videogenic identifies the officiant address of the wedding, the skateboard kickflip, the weightlifter completing the clean and jerk, the bird carrying its prey, the graduation hat toss, and the total solar eclipse. In addition, Videogenic inherits knowledge on good composition and framing from professional photographs (e.g., low-angle shot for skateboarding (Figure \ref{fig:human-evaluation-examples}b) and close-up shot of the hunting bird (Figure \ref{fig:human-evaluation-examples}d)).

% \subsubsection{Failure Cases}
% Since Videogenic uses static photographs as its prior, it does not make use of motion-related information. Thus, we see that some highlights, while capturing good moments, are a little monotonous. In these cases, participants may have preferred the baseline highlight videos that had more movement. We discuss how this may be addressed in Section \ref{section:future-work}.

\section{Expert Study}
We conduct a within-subjects expert study to evaluate the usefulness of Videogenic in helping video editors create highlight videos. The following outlines our study design, participants, procedure, and results. Our main research questions are:

\begin{itemize}
    \item [RQ1.] How would participants' workload level be affected with the use of Videogenic?
    \item [RQ2.] How do participants' find the usability of Videogenic?
    \item [RQ3.] How would the use of Videogenic affect the participants' task completion time?
    \item [RQ4.] Qualitatively, what would participants see as the pros and cons of Videogenic?
\end{itemize}

\subsection{Study Design}

\subsubsection{Independent Variable}
The independent variable of the study is the system: Videogenic versus a baseline of manual editing. In the experimental condition, we ask participants to create a highlight video using Videogenic. In the baseline condition, we ask participants to create a highlight video using Adobe Premiere Pro, a standard video editing software that editors use to create highlight videos.

\subsubsection{Dependent Variable}
The dependent variables of the study are workload (RQ1) measured by the mental, temporal, effort, and frustration components of the NASA TLX questionnaire \cite{nasa-tlx}, usability (RQ2) measured by the System Usability Scale (SUS) questionnaire \cite{sus}, and task completion time (RQ3) reported by the participant (in seconds). All questionnaire questions are represented on a 7-point Likert scale.

\begin{figure*}[tbp]
  \centering
  \includegraphics[width=15cm]{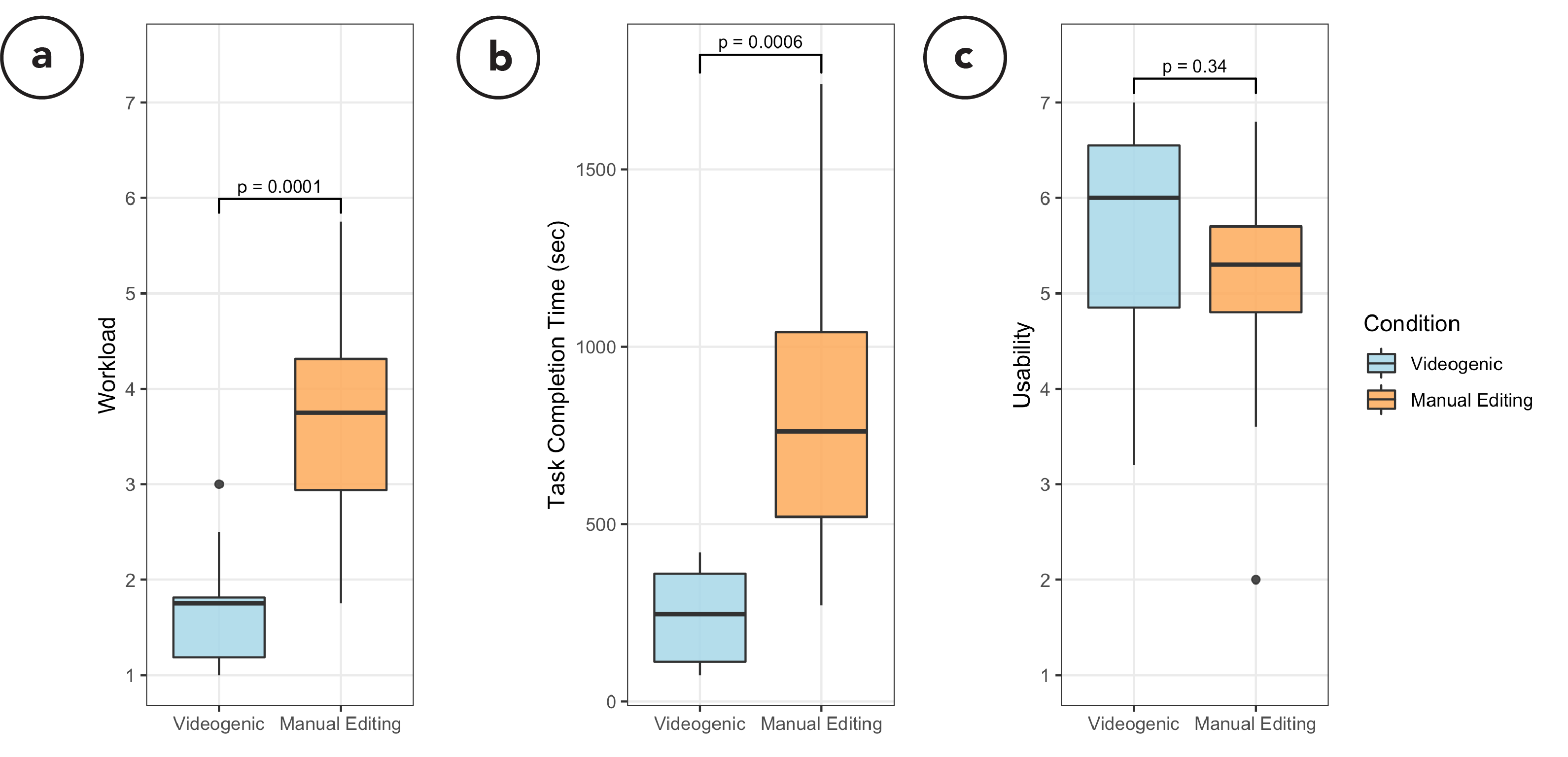}
  \caption{Expert study results ($N$=12). Boxplots from left to right: workload measured with NASA TLX \cite{nasa-tlx} (7-point Likert scale, lower is better) (a), task completion time (seconds, lower is better) (b), and usability measured with SUS \cite{sus} (7-point Likert scale, higher is better) (c). The baseline is manual editing with Adobe Premiere Pro.}
  \Description{Expert study results (N=12). Boxplots from left to right: workload measured with NASA TLX (7-point Likert scale, lower is better) (a), task completion time (seconds, lower is better) (b), and usability measured with SUS (7-point Likert scale, higher is better) (c). The baseline is manual editing with Adobe Premiere Pro.}
  \label{fig:expert-study-results}
\end{figure*}

\subsection{Participants}
We recruit 12 professional video editors (4 female, 8 male) aged 18 to 48 (mean=28.75, SD=9.77) from Upwork \cite{upwork}, a platform for hiring freelancers. We conduct a background survey with the participants before each study to assess their video editing experience. Overall, participants have high self-rated familiarity with video editing (mean=6.25, SD=0.75) (7-point Likert scale) and have several years of experience (mean=6.71, SD=3.93). All participants regularly use Adobe Premiere Pro for video editing.
% In addition, participants also have experience with other video editing software such as Adobe After Effects, Final Cut Pro, DaVinci Resolve, and Sony Vegas Pro.

\subsection{Procedure}
We conduct the expert study remotely. After receiving the participant's consent, we collect information about individual backgrounds. We then ask the participant to create a highlight video from a source video with Videogenic and to create a highlight video from another source video by manually editing in Adobe Premiere Pro. We counterbalance both the order of the conditions and the order of the source videos. The source videos are of comparable difficulty, both being videos containing a large variety of complex scenes that depict a full-day vlog of an activity (i.e., skydiving and surfing). We also ask participants to record the time they spend in each condition by starting a timer after opening the application and stopping the timer after completing the highlight video.
% This includes rendering time for both Videogenic and manual editing, which are both relatively small considering that the final highlight videos are short in length}.
After each condition, we ask participants to complete the NASA TLX, SUS, and task completion time questionnaires. After the participant completes both conditions, we ask the participant to answer open-ended questions regarding the overall experience of using Videogenic. The study lasts for approximately 40 minutes. We compensate participants \$30 USD for their time.

\begin{figure*}[tbp]
  \centering
  \includegraphics[width=9.5cm]{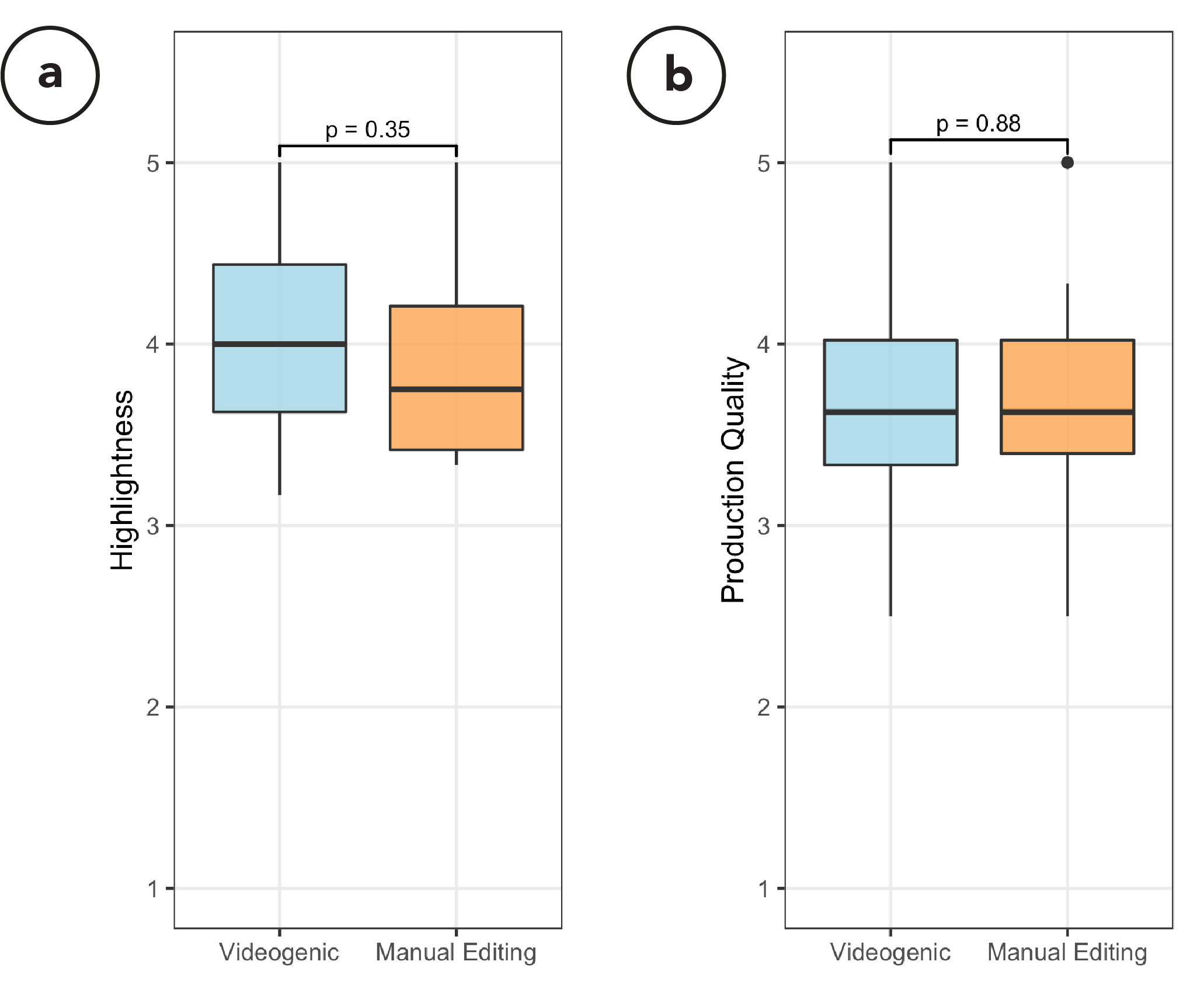}
  \caption{Highlightness and quality results ($N$=20). Boxplots from left to right: highlightness (5-point Likert scale, higher is better) (a), production quality (5-point Likert scale, higher is better) (b). The baseline is manual editing with Adobe Premiere Pro.}
  \Description{Highlightness and quality results (N=20). Boxplots from left to right: highlightness (5-point Likert scale, higher is better) (a), production quality (5-point Likert scale, higher is better) (b). The baseline is manual editing with Adobe Premiere Pro.}
  \label{fig:quality-evaluation-results}
\end{figure*}

\subsection{Results and Discussion}
For quantitative analysis, we analyze the scores for workload, usability, and task completion through paired t-tests.
In addition, we recruit external raters to evaluate whether the final highlight videos created by the participants capture the highlight moments of activities and have good production quality. We analyze these external ratings through unpaired t-tests.
Figure \ref{fig:expert-study-results} shows an overview of the quantitative results comparing Videogenic against the manual editing baseline.
For qualitative analysis, we analyze the participants' open-ended responses with deductive thematic analysis \cite{thematic-analysis} according to the dimensions of the quantitative measurements (i.e., workload, task completion time, and usability).
% The following presents the results of the expert study \rev{and discusses potential implications for video editing work, such as reducing tedious `chores', motivating content creators to create more video content, and inspiring new interfaces and workflows for highlight video creation.}

\subsubsection{Workload}
The differences in workload per participant across conditions pass the Shapiro-Wilk test of normality ($W$=0.95, $p$=0.60). We thus compare the differences in workload through a parametric paired t-test. Participants report a significantly lower workload when using Videogenic (mean=1.69, SD=0.61) compared to the baseline (mean=3.67, SD=1.20) ($t$(11)=5.84, $p$=0.0001, $r$=0.87, $d_s$=1.68) (7-point Likert scale, lower is better) (Figure \ref{fig:expert-study-results}a). Participants state that "\textit{[Videogenic] found the most interesting moments for the clip} (P11)" and that it "\textit{does away a lot of the bland and monotonous editing `chores' like having to scrub through a lot of fluff} (P2)". Participants enjoy the flexibility of multiple modes: "\textit{In auto mode, it is literally zero effort. In user selection mode, the fact that [Videogenic] gets you 90\% there is pretty cool too.} (P2)" Overall, participants feel that Videogenic helps reduce the tedious components of creating highlight videos. For example, in action footage such as skateboarding or surfing, the camera has to be constantly rolling to capture unexpected moments. However, going through the raw footage can be a tiresome chore. 
% By reducing the ``chore'' of finding highlight moments, 
Videogenic could change the nature of editing work by allowing editors to dedicate more mental energy to the creative aspects of editing. For content creators, this could motivate them to create and share more video content.

\subsubsection{Task Completion Time}
The differences in task completion time per participant across conditions pass the Shapiro-Wilk test of normality ($W$=0.93, $p$=0.36). We thus compare the differences in task completion time through a parametric paired t-test. Participants report a significantly lower task completion time in seconds when using Videogenic (mean=247, SD=132) compared to the baseline (mean=856, SD=476) ($t$(11)=4.73, $p$=0.0006, $r$=0.82, $d_s$=1.37) (time taken in seconds, lower is better) (Figure \ref{fig:expert-study-results}b). Participants state that Videogenic helps "\textit{cut the time when it comes to searching for the right clips to use} (P1)": "\textit{With manual editing, I have to sort through a 5-minute video just for that 5 seconds of good footage.} (P1)" Several participants mention how Videogenic is especially suitable for editing short videos: "\textit{Premiere and other software can be too clunky to make short videos} (P8)" and "\textit{no matter how short the project is, [Premiere] always takes more time than I would like it to} (P6)". One participant also suggests a combination of the two systems: "\textit{I'd use [Videogenic] to detect highlight moments for further editing in Premiere} (P10)". Overall, participants feel that Videogenic significantly shortens the time it takes to create a highlight video, notably by reducing the time spent on searching for highlight moments within large amounts of footage.
% For example, a video production workflow in a day of filming may involve tens of video clips amounting to hours of raw footage that an editor must sift through to surface the handful of highlight moments.
% The professional work time of editors is valuable and time saved by Videogenic could mean more time to create higher quality edits. For non-professional video content creators, Videogenic could allow creators to allocate more time into other stages of the content creation process, such as crafting an engaging story.}

\subsubsection{Usability}
The differences in usability per participant across conditions pass the Shapiro-Wilk test of normality ($W$=0.95, $p$=0.64). We thus compare the differences in usability through a parametric paired t-test. Participants report a higher but not significantly higher usability when using Videogenic (mean=5.62, SD=1.34) compared to the baseline (mean=5.05, SD=1.29) ($t$(11)=-0.99, $p$=0.34) (7-point Likert scale, higher is better) (Figure \ref{fig:expert-study-results}c). The absence of statistical significance is unsurprising as the baseline (Adobe Premiere Pro) is a polished editing software that the participants are familiar with. Participants found Videogenic to be "\textit{a very simple feature that works well and is easy to navigate and use} (P6)" and is "\textit{pretty fool-proof} (P7)": "\textit{I've never used the program before, but was still able to create the video I needed in under two minutes. That's incredible.} (P6)".

Participants also comment on specific components of Videogenic:

\textbf{Highlight Graph.} Participants enjoy scrubbing through the highlight graph "\textit{to see spots where there could be highlights in graph form}" (P8): "\textit{It shows me the high and low points of the video and where to cut.} (P1)" Participants also appreciate being able to see the data: "\textit{it made [Videogenic] feel sophisticated because of the data being shown} (P7)".
% \textbf{Highlight Template.} Participants express that "\textit{the effects and songs used on Videogenic were catchy and made for a cool video} (P8)" and appreciate the multiple stylistic options ("\textit{I like that [Videogenic] had [different] styles of video it would produce when you pick the song [such as] `hype' and other [genre] keywords}" (P7)).

Overall, participants express that Videogenic is easy to use to create highlight videos. Given that many editors enjoy interacting with the highlight graph, it could potentially be useful for the highlight graph to be integrated within video editing programs.
% Such an interface could be a pro-level tool for editors to visualize the distribution of highlight moments across the video and have manual control over which segments to use for the final highlight clip.
% To also show the potential for more automatic tools, the author created several example highlight templates. Future developers may also create new and more creative highlight templates built on Videogenic. For content creators, the highlight template could be an easy way of achieving stylized highlight videos.

\subsubsection{Quality}
We recruit \camera{external} raters to evaluate the quality of the final highlight videos created by the editors using Videogenic and using the baseline.

\textbf{Procedure.}
We recruit 20 US-based raters on Prolific \cite{prolific} with standard sampling and prescreen participants such that they must have experience in using TikTok so that they are familiar with the concept of highlight videos. After receiving participants' consent, we ask each participant to rate the quality of 24 highlight videos (12 created with Videogenic, 12 created with baseline) by answering how they feel about the following two statements on a scale of 1 to 5 (strongly disagree, disagree, neutral, agree, strongly agree):

\begin{itemize}
\item {The video captures the highlight moments of <activity>.}
\item{This is a well-made highlight video.}
\end{itemize}

% The participant does not know which highlight video is created using Videogenic or are manually edited and we randomize the videos' orderings in the questionnaire. To ensure the quality of the responses, we have a check to make sure participants are able to see the videos correctly and we include two quality control questions \cite{agley2022quality}.
The study takes approximately 10 minutes to complete and we compensate participants \$2 USD for their time.

\textbf{Results.}
Figure \ref{fig:quality-evaluation-results} shows an overview of the results.
The differences per participant across conditions pass the Shapiro-Wilk test of normality for both "highlightness" ($W$=0.98, $p$=0.93) and production quality ($W$=0.93, $p$=0.16). We thus analyze the results for statistical significance through parametric unpaired t-tests. There is no significant statistical difference between the highlight videos created using Videogenic compared to the baseline (manual editing with Premiere) for both highlightness and production quality. Participants rate a slightly higher highlightness for the videos created using Videogenic (mean=4.06, SD=0.59) compared to the baseline (mean=3.89, SD=0.55) ($t$(37.9)=0.95, $p$=0.35) (5-point Likert scale, higher is better) (Figure \ref{fig:quality-evaluation-results}a). Participants rate a similar production quality for the videos created using Videogenic (mean=3.70, SD=0.66) compared to the baseline (mean=3.73, SD=0.60) ($t$(37.7)=-0.15, $p$=0.88) (5-point Likert scale, higher is better) (Figure \ref{fig:quality-evaluation-results}b).
Overall, raters feel that both videos created by Videogenic and by human editors capture the highlight moments of activities and are well-made as rated on our 5-point scale.
% Raters feel that the videos created by Videogenic capture the highlight moments of activities even slightly better than those created by human editors. Raters feel that production quality of the videos created by Videogenic are on par with those created by human editors.}

\begin{figure*}[!htb]
  \centering
  \includegraphics[width=15cm]{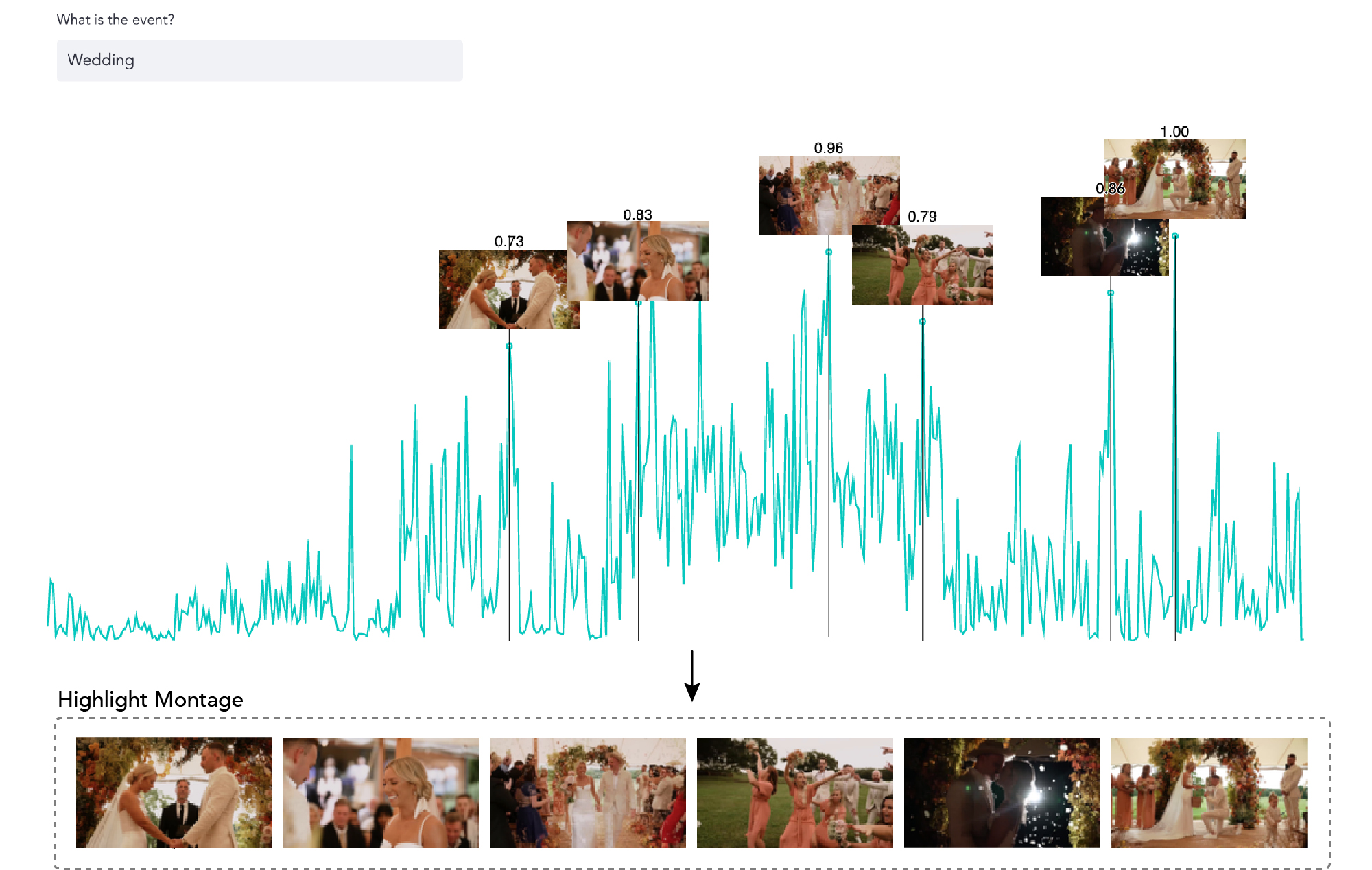}
  \caption{\rev{Videogenic can be extended to create a highlight moments montage by combining multiple local maxima highlight moments.}}
  \Description{Videogenic can be extended to create a highlight moments montage by combining multiple local maxima highlight moments.}
  \label{fig:highlight-montage}
\end{figure*}

\section{Extended Applications}

\rev{We implemented two extended applications to explore video applications that could use Videogenic as a building block.}

\begin{figure*}[!htb]
  \centering
  \includegraphics[width=15cm]{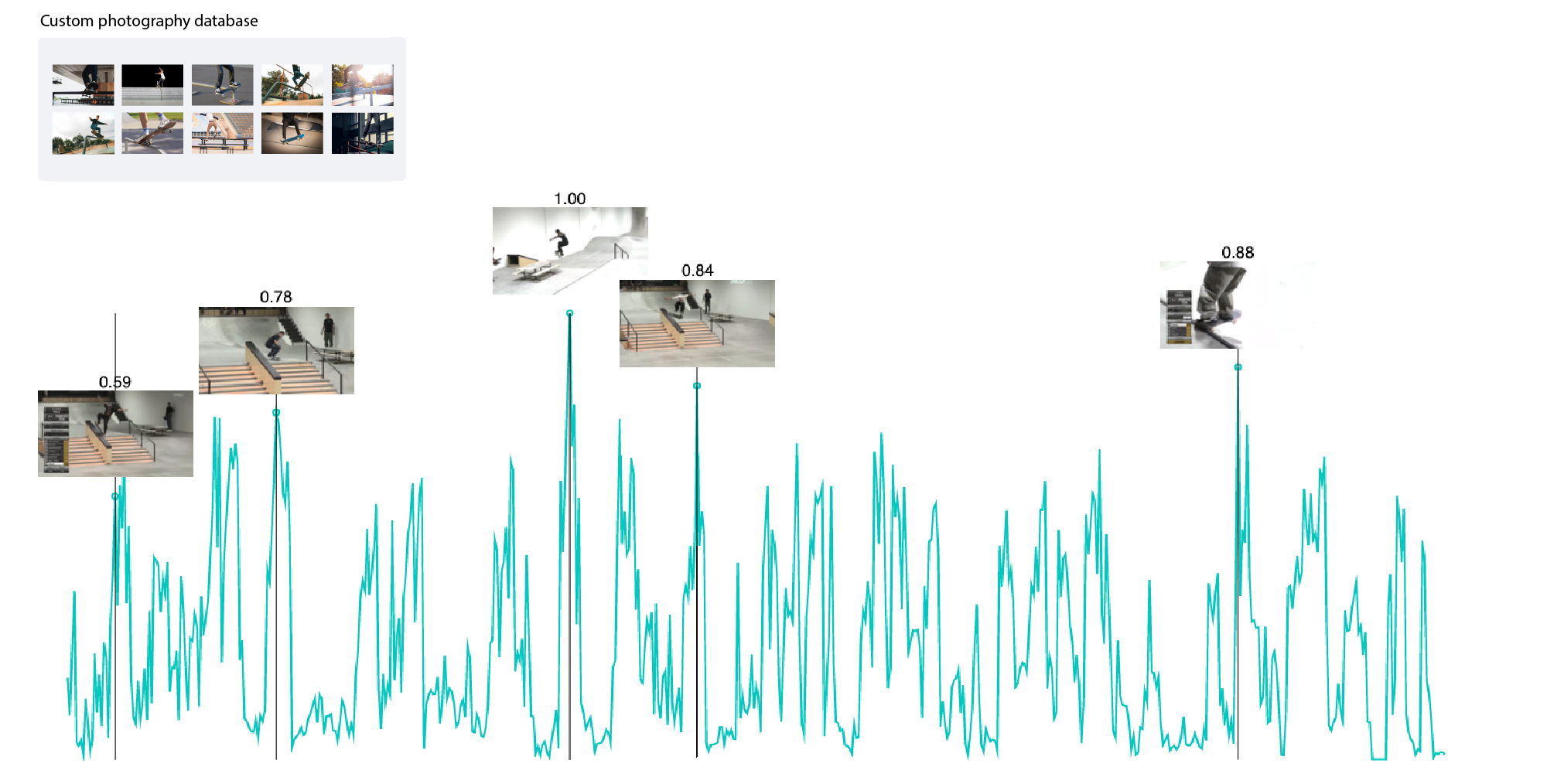}
  \caption{\rev{Videogenic can be extended to create a personalized highlight video by creating a custom photography database.}}
  \Description{Videogenic can be extended to create a personalized highlight video by creating a custom photography database.}
  \label{fig:personalization}
\end{figure*}

\subsection{Highlight Moments Montage}

\rev{A highlight moments montage features a sequence of many highlight moments from an event. Instead of identifying one key highlight moment, we may extend Videogenic to identify multiple highlight moments by computing several local maxima. Figure \ref{fig:highlight-montage} shows how several different local highlight moments from a wedding video can be combined to create a wedding highlight moments montage.}

\subsection{Personalized Highlight Video}
\label{section:personalized-database}

\rev{Videogenic can be easily adapted to support personalized highlight videos. The user may upload their own photographs to create a custom photography database, powering a personalized Videogenic. Figure \ref{fig:personalization} shows how a photography database containing a particular skateboarding trick can help create a personalized highlight video featuring this trick.}

\section{\rev{Limitations and Future Work}}
\label{section:future-work}
While Videogenic was positively received in our user studies, there are several avenues for improvement that we plan to address for future work.
\camera{First, Videogenic determines highlights from the visual domain. Thus, Videogenic is not designed for specific categories of audio-oriented videos with few visual concepts, such as recordings of podcasts or interview videos. In addition, further work on gaining a deeper understanding of the target domain may be required to extend Videogenic for videos involving complex subtasks, such as how-to videos \cite{chi2013democut, truong2021automatic, chang2021rubyslippers}.}
Second, as Videogenic uses photographs as its prior, it currently does not factor in motion information from videos. \rev{We hypothesize this to be the main reason why the highlight videos automatically generated with Videogenic for surfing and skydiving received lower preference (Figure \ref{fig:human-evaluation-results}), as they were more static shots. While this may be addressed through human-in-the-loop selection of highlight segments (Section \ref{section:user-selection}), we may extend Videogenic to additionally leverage \textit{professional stock videos} as a prior. We could represent motion information using video motion extraction methods such as optical flow \cite{lin2020introduction}}.
\rev{Third, Videogenic uses a small collection of stock photographs to compute highlight scores. There may be cases where novel activities do not have readily available professional photographs. As the number of photographs required by Videogenic is small (10 photographs), we could allow users to create their own personalized examples (Section \ref{section:personalized-database}).} \camera{A personalized example set could be an interesting form of control for users to steer the direction of the highlight moments from a handful of personal examples \cite{pseudoclient}.
Finally, Videogenic currently takes in an average photo representation from a collection of photographs. While we found this to generally perform well through extensive human evaluations, additional experiments could be done, such as ablating different numbers of photographs for different video domains or having multiple clustered averages instead of a single average (e.g., photograph clusters of handshakes, crowd shots, and thrown hats for a "graduation" highlight video). This could also enable greater a level of customization of highlight videos.}

\section{Conclusion}
In this paper, we present Videogenic, a \rev{simple yet very effective technique} for creating highlight videos. Our key insight is that professional photographs tend to capture the most remarkable moments of a given activity. We conduct a human evaluation study ($N$=50), showing that a set of high-quality photographs combined with encodings of CLIP can act as a strong prior for extracting domain-specific highlights for videos encompassing a diverse range of domains. We further evaluate the usefulness of Videogenic for video editors through a within-subjects expert study ($N$=12) comparing Videogenic to a baseline (Adobe Premiere Pro), demonstrating decreased workload, decreased task completion time, and increased usability. \rev{External raters rated high "highlightness" and production quality for the highlight videos created by editors with Videogenic.
This work takes a step towards \textit{out-of-the-box}, domain-agnostic highlight video generation by building on the domain knowledge of photographers. In recent years, we see growth in long-form video content (e.g., livestreaming \cite{livestreaming-growth}) as well as a proliferation of video capturing devices (e.g., smartphones and action cameras \cite{insta360}). On the other hand, we see a rapid surge in popularity in short-form video consumption (e.g., TikTok, Instagram Reels, and YouTube Shorts). We hope Videogenic can help to bridge this gap by lowering the barrier required to convert long-form videos into engaging short-form highlights.}

% \raggedbottom
% \pagebreak


\begin{acks}
\rev{We would like to thank colleagues in the Augmented Design Capability Studio for providing valuable feedback on our system.}
\end{acks}

\bibliographystyle{ACM-Reference-Format}
\bibliography{main}


\begin{thebibliography}{44}

%%% the Web address.

\ifx \showCODEN    \undefined \def \showCODEN     #1{\unskip}     \fi
\ifx \showDOI      \undefined \def \showDOI       #1{#1}\fi
\ifx \showISBNx    \undefined \def \showISBNx     #1{\unskip}     \fi
\ifx \showISBNxiii \undefined \def \showISBNxiii  #1{\unskip}     \fi
\ifx \showISSN     \undefined \def \showISSN      #1{\unskip}     \fi
\ifx \showLCCN     \undefined \def \showLCCN      #1{\unskip}     \fi
\ifx \shownote     \undefined \def \shownote      #1{#1}          \fi
\ifx \showarticletitle \undefined \def \showarticletitle #1{#1}   \fi
\ifx \showURL      \undefined \def \showURL       {\relax}        \fi
% invisible to TeX
\providecommand\bibfield[2]{#2}
\providecommand\bibinfo[2]{#2}
\providecommand\natexlab[1]{#1}
\providecommand\showeprint[2][]{arXiv:#2}

\bibitem[sho(2021)]%
        {short-video-app}
 \bibinfo{year}{2021}\natexlab{}.
\newblock \bibinfo{booktitle}{\emph{Apple announces the 2021 App Store Award
  winners and most downloaded apps of the year}}.
\newblock
\urldef\tempurl%
\url{https://techcrunch.com/2021/12/02/apple-announces-the-2021-app-store-award-winners-and-most-downloaded-
  apps-of-the-year/}
\showURL{%
Retrieved August 15, 2022 from \tempurl}


\bibitem[fla(2022)]%
        {flashcut}
 \bibinfo{year}{2022}\natexlab{}.
\newblock \bibinfo{booktitle}{\emph{FlashCut Auto Editing - ONE R Support}}.
\newblock
\urldef\tempurl%
\url{https://onlinemanual.insta360.com/oner/en-us/editing/flashcut}
\showURL{%
Retrieved August 15, 2022 from \tempurl}


\bibitem[gop(2022)]%
        {gopro}
 \bibinfo{year}{2022}\natexlab{}.
\newblock \bibinfo{booktitle}{\emph{HERO10 Black 5.3K Video 23MP Action Camera
  Bundle | GoPro}}.
\newblock
\urldef\tempurl%
\url{https://gopro.com/en/us/shop/cameras/hero10-black/CHDHX-101-master.html}
\showURL{%
Retrieved August 15, 2022 from \tempurl}


\bibitem[ins(2022)]%
        {insta360}
 \bibinfo{year}{2022}\natexlab{}.
\newblock \bibinfo{booktitle}{\emph{Insta360 ONE RS – Waterproof Action
  Camera + 360 Camera in One}}.
\newblock
\urldef\tempurl%
\url{https://www.insta360.com/product/insta360-oners}
\showURL{%
Retrieved August 15, 2022 from \tempurl}


\bibitem[liv(2022)]%
        {livestreaming-growth}
 \bibinfo{year}{2022}\natexlab{}.
\newblock \bibinfo{booktitle}{\emph{Live Streaming Market Worth \$4.26 Billion
  by 2028}}.
\newblock
\urldef\tempurl%
\url{https://www.bloomberg.com/press-releases/2022-05-05/live-streaming-market-worth-4-26-billion-by-2028-market-size-share-forecasts-trends-analysis-report-with-covid-19-impact}
\showURL{%
Retrieved August 15, 2022 from \tempurl}


\bibitem[pro(2022)]%
        {prolific}
 \bibinfo{year}{2022}\natexlab{}.
\newblock \bibinfo{booktitle}{\emph{Prolific}}.
\newblock
\urldef\tempurl%
\url{https://www.prolific.co/}
\showURL{%
Retrieved August 15, 2022 from \tempurl}


\bibitem[upw(2022)]%
        {upwork}
 \bibinfo{year}{2022}\natexlab{}.
\newblock \bibinfo{booktitle}{\emph{Upwork}}.
\newblock
\urldef\tempurl%
\url{https://www.upwork.com/}
\showURL{%
Retrieved August 15, 2022 from \tempurl}


\bibitem[whi(2022)]%
        {which-frame}
 \bibinfo{year}{2022}\natexlab{}.
\newblock \bibinfo{booktitle}{\emph{Which Frame?}}
\newblock
\urldef\tempurl%
\url{http://whichfra.me/}
\showURL{%
Retrieved August 15, 2022 from \tempurl}


\bibitem[you(2022)]%
        {youtube-preview}
 \bibinfo{year}{2022}\natexlab{}.
\newblock \bibinfo{booktitle}{\emph{YouTube Help - Video previews}}.
\newblock
\urldef\tempurl%
\url{https://support.google.com/youtube/answer/7074781?hl=en}
\showURL{%
Retrieved August 15, 2022 from \tempurl}


\bibitem[Amershi et~al\mbox{.}(2019)]%
        {amershi2019guidelines}
\bibfield{author}{\bibinfo{person}{Saleema Amershi}, \bibinfo{person}{Dan
  Weld}, \bibinfo{person}{Mihaela Vorvoreanu}, \bibinfo{person}{Adam Fourney},
  \bibinfo{person}{Besmira Nushi}, \bibinfo{person}{Penny Collisson},
  \bibinfo{person}{Jina Suh}, \bibinfo{person}{Shamsi Iqbal},
  \bibinfo{person}{Paul~N Bennett}, \bibinfo{person}{Kori Inkpen},
  {et~al\mbox{.}}} \bibinfo{year}{2019}\natexlab{}.
\newblock \showarticletitle{Guidelines for human-AI interaction}. In
  \bibinfo{booktitle}{\emph{Proceedings of the 2019 chi conference on human
  factors in computing systems}}. \bibinfo{pages}{1--13}.
\newblock


\bibitem[Bernstein et~al\mbox{.}(2011)]%
        {bernstein2011crowds}
\bibfield{author}{\bibinfo{person}{Michael~S Bernstein}, \bibinfo{person}{Joel
  Brandt}, \bibinfo{person}{Robert~C Miller}, {and} \bibinfo{person}{David~R
  Karger}.} \bibinfo{year}{2011}\natexlab{}.
\newblock \showarticletitle{Crowds in two seconds: Enabling realtime
  crowd-powered interfaces}. In \bibinfo{booktitle}{\emph{Proceedings of the
  24th annual ACM symposium on User interface software and technology}}.
  \bibinfo{pages}{33--42}.
\newblock


\bibitem[Braun and Clarke(2006)]%
        {thematic-analysis}
\bibfield{author}{\bibinfo{person}{Virginia Braun} {and}
  \bibinfo{person}{Victoria Clarke}.} \bibinfo{year}{2006}\natexlab{}.
\newblock \showarticletitle{Using thematic analysis in psychology}.
\newblock \bibinfo{journal}{\emph{Qualitative research in psychology}}
  \bibinfo{volume}{3}, \bibinfo{number}{2} (\bibinfo{year}{2006}),
  \bibinfo{pages}{77--101}.
\newblock


\bibitem[Brooke et~al\mbox{.}(1996)]%
        {sus}
\bibfield{author}{\bibinfo{person}{John Brooke} {et~al\mbox{.}}}
  \bibinfo{year}{1996}\natexlab{}.
\newblock \showarticletitle{SUS-A quick and dirty usability scale}.
\newblock \bibinfo{journal}{\emph{Usability evaluation in industry}}
  \bibinfo{volume}{189}, \bibinfo{number}{194} (\bibinfo{year}{1996}),
  \bibinfo{pages}{4--7}.
\newblock


\bibitem[Chang et~al\mbox{.}(2021)]%
        {chang2021rubyslippers}
\bibfield{author}{\bibinfo{person}{Minsuk Chang}, \bibinfo{person}{Mina Huh},
  {and} \bibinfo{person}{Juho Kim}.} \bibinfo{year}{2021}\natexlab{}.
\newblock \showarticletitle{Rubyslippers: Supporting content-based voice
  navigation for how-to videos}. In \bibinfo{booktitle}{\emph{Proceedings of
  the 2021 CHI conference on human factors in computing systems}}.
  \bibinfo{pages}{1--14}.
\newblock


\bibitem[Chi et~al\mbox{.}(2013)]%
        {chi2013democut}
\bibfield{author}{\bibinfo{person}{Pei-Yu Chi}, \bibinfo{person}{Joyce Liu},
  \bibinfo{person}{Jason Linder}, \bibinfo{person}{Mira Dontcheva},
  \bibinfo{person}{Wilmot Li}, {and} \bibinfo{person}{Bjoern Hartmann}.}
  \bibinfo{year}{2013}\natexlab{}.
\newblock \showarticletitle{Democut: generating concise instructional videos
  for physical demonstrations}. In \bibinfo{booktitle}{\emph{Proceedings of the
  26th annual ACM symposium on User interface software and technology}}.
  \bibinfo{pages}{141--150}.
\newblock


\bibitem[Chu et~al\mbox{.}(2015)]%
        {chu2015video}
\bibfield{author}{\bibinfo{person}{Wen-Sheng Chu}, \bibinfo{person}{Yale Song},
  {and} \bibinfo{person}{Alejandro Jaimes}.} \bibinfo{year}{2015}\natexlab{}.
\newblock \showarticletitle{Video co-summarization: Video summarization by
  visual co-occurrence}. In \bibinfo{booktitle}{\emph{Proceedings of the IEEE
  conference on computer vision and pattern recognition}}.
  \bibinfo{pages}{3584--3592}.
\newblock


\bibitem[Fechner(1860)]%
        {2afc}
\bibfield{author}{\bibinfo{person}{Gustav~Theodor Fechner}.}
  \bibinfo{year}{1860}\natexlab{}.
\newblock \bibinfo{booktitle}{\emph{Elemente der psychophysik}}.
  Vol.~\bibinfo{volume}{2}.
\newblock \bibinfo{publisher}{Breitkopf u. H{\"a}rtel}.
\newblock


\bibitem[Gygli et~al\mbox{.}(2016)]%
        {gygli2016video2gif}
\bibfield{author}{\bibinfo{person}{Michael Gygli}, \bibinfo{person}{Yale Song},
  {and} \bibinfo{person}{Liangliang Cao}.} \bibinfo{year}{2016}\natexlab{}.
\newblock \showarticletitle{Video2gif: Automatic generation of animated gifs
  from video}. In \bibinfo{booktitle}{\emph{Proceedings of the IEEE conference
  on computer vision and pattern recognition}}. \bibinfo{pages}{1001--1009}.
\newblock


\bibitem[Hart and Staveland(1988)]%
        {nasa-tlx}
\bibfield{author}{\bibinfo{person}{Sandra~G Hart} {and}
  \bibinfo{person}{Lowell~E Staveland}.} \bibinfo{year}{1988}\natexlab{}.
\newblock \showarticletitle{Development of NASA-TLX (Task Load Index): Results
  of empirical and theoretical research}.
\newblock In \bibinfo{booktitle}{\emph{Advances in psychology}}.
  Vol.~\bibinfo{volume}{52}. \bibinfo{publisher}{Elsevier},
  \bibinfo{pages}{139--183}.
\newblock


\bibitem[Khosla et~al\mbox{.}(2013)]%
        {khosla2013large}
\bibfield{author}{\bibinfo{person}{Aditya Khosla}, \bibinfo{person}{Raffay
  Hamid}, \bibinfo{person}{Chih-Jen Lin}, {and} \bibinfo{person}{Neel
  Sundaresan}.} \bibinfo{year}{2013}\natexlab{}.
\newblock \showarticletitle{Large-scale video summarization using web-image
  priors}. In \bibinfo{booktitle}{\emph{Proceedings of the IEEE conference on
  computer vision and pattern recognition}}. \bibinfo{pages}{2698--2705}.
\newblock


\bibitem[Kim et~al\mbox{.}(2018)]%
        {kim2018exploiting}
\bibfield{author}{\bibinfo{person}{Hoseong Kim}, \bibinfo{person}{Tao Mei},
  \bibinfo{person}{Hyeran Byun}, {and} \bibinfo{person}{Ting Yao}.}
  \bibinfo{year}{2018}\natexlab{}.
\newblock \showarticletitle{Exploiting web images for video highlight detection
  with triplet deep ranking}.
\newblock \bibinfo{journal}{\emph{IEEE Transactions on Multimedia}}
  \bibinfo{volume}{20}, \bibinfo{number}{9} (\bibinfo{year}{2018}),
  \bibinfo{pages}{2415--2426}.
\newblock


\bibitem[Kolekar and Sengupta(2006)]%
        {kolekar2006event}
\bibfield{author}{\bibinfo{person}{Maheshkumar~H Kolekar} {and}
  \bibinfo{person}{Somnath Sengupta}.} \bibinfo{year}{2006}\natexlab{}.
\newblock \showarticletitle{Event-importance based customized and automatic
  cricket highlight generation}. In \bibinfo{booktitle}{\emph{2006 IEEE
  International Conference on Multimedia and Expo}}. IEEE,
  \bibinfo{pages}{1617--1620}.
\newblock


\bibitem[Krause et~al\mbox{.}(2016)]%
        {krause2016interacting}
\bibfield{author}{\bibinfo{person}{Josua Krause}, \bibinfo{person}{Adam Perer},
  {and} \bibinfo{person}{Kenney Ng}.} \bibinfo{year}{2016}\natexlab{}.
\newblock \showarticletitle{Interacting with predictions: Visual inspection of
  black-box machine learning models}. In \bibinfo{booktitle}{\emph{Proceedings
  of the 2016 CHI conference on human factors in computing systems}}.
  \bibinfo{pages}{5686--5697}.
\newblock


\bibitem[Lee et~al\mbox{.}(2012)]%
        {lee2012discovering}
\bibfield{author}{\bibinfo{person}{Yong~Jae Lee}, \bibinfo{person}{Joydeep
  Ghosh}, {and} \bibinfo{person}{Kristen Grauman}.}
  \bibinfo{year}{2012}\natexlab{}.
\newblock \showarticletitle{Discovering important people and objects for
  egocentric video summarization}. In \bibinfo{booktitle}{\emph{2012 IEEE
  conference on computer vision and pattern recognition}}. IEEE,
  \bibinfo{pages}{1346--1353}.
\newblock


\bibitem[Lin(2020)]%
        {lin2020introduction}
\bibfield{author}{\bibinfo{person}{Chuan-en Lin}.}
  \bibinfo{year}{2020}\natexlab{}.
\newblock \showarticletitle{Introduction to motion estimation with optical
  flow}.
\newblock \bibinfo{journal}{\emph{can be found under https://nanonets.
  com/blog/optical-flow}} (\bibinfo{year}{2020}).
\newblock


\bibitem[Lin and Martelaro(2021)]%
        {pseudoclient}
\bibfield{author}{\bibinfo{person}{David Chuan-En Lin} {and}
  \bibinfo{person}{Nikolas Martelaro}.} \bibinfo{year}{2021}\natexlab{}.
\newblock \showarticletitle{Learning Personal Style from Few Examples}. In
  \bibinfo{booktitle}{\emph{Designing Interactive Systems Conference 2021}}.
  \bibinfo{pages}{1566--1578}.
\newblock


\bibitem[Lu and Grauman(2013)]%
        {lu2013story}
\bibfield{author}{\bibinfo{person}{Zheng Lu} {and} \bibinfo{person}{Kristen
  Grauman}.} \bibinfo{year}{2013}\natexlab{}.
\newblock \showarticletitle{Story-driven summarization for egocentric video}.
  In \bibinfo{booktitle}{\emph{Proceedings of the IEEE conference on computer
  vision and pattern recognition}}. \bibinfo{pages}{2714--2721}.
\newblock


\bibitem[Matejka et~al\mbox{.}(2013)]%
        {matejka2013swifter}
\bibfield{author}{\bibinfo{person}{Justin Matejka}, \bibinfo{person}{Tovi
  Grossman}, {and} \bibinfo{person}{George Fitzmaurice}.}
  \bibinfo{year}{2013}\natexlab{}.
\newblock \showarticletitle{Swifter: improved online video scrubbing}. In
  \bibinfo{booktitle}{\emph{Proceedings of the SIGCHI Conference on Human
  Factors in Computing Systems}}. \bibinfo{pages}{1159--1168}.
\newblock


\bibitem[Matejka et~al\mbox{.}(2014)]%
        {matejka2014video}
\bibfield{author}{\bibinfo{person}{Justin Matejka}, \bibinfo{person}{Tovi
  Grossman}, {and} \bibinfo{person}{George Fitzmaurice}.}
  \bibinfo{year}{2014}\natexlab{}.
\newblock \showarticletitle{Video lens: rapid playback and exploration of large
  video collections and associated metadata}. In
  \bibinfo{booktitle}{\emph{Proceedings of the 27th annual ACM symposium on
  User interface software and technology}}. \bibinfo{pages}{541--550}.
\newblock


\bibitem[Nepal et~al\mbox{.}(2001)]%
        {nepal2001automatic}
\bibfield{author}{\bibinfo{person}{Surya Nepal}, \bibinfo{person}{Uma
  Srinivasan}, {and} \bibinfo{person}{Graham Reynolds}.}
  \bibinfo{year}{2001}\natexlab{}.
\newblock \showarticletitle{Automatic detection of'Goal'segments in basketball
  videos}. In \bibinfo{booktitle}{\emph{Proceedings of the ninth ACM
  international conference on Multimedia}}. \bibinfo{pages}{261--269}.
\newblock


\bibitem[Radford et~al\mbox{.}(2021)]%
        {clip}
\bibfield{author}{\bibinfo{person}{Alec Radford}, \bibinfo{person}{Jong~Wook
  Kim}, \bibinfo{person}{Chris Hallacy}, \bibinfo{person}{Aditya Ramesh},
  \bibinfo{person}{Gabriel Goh}, \bibinfo{person}{Sandhini Agarwal},
  \bibinfo{person}{Girish Sastry}, \bibinfo{person}{Amanda Askell},
  \bibinfo{person}{Pamela Mishkin}, \bibinfo{person}{Jack Clark},
  {et~al\mbox{.}}} \bibinfo{year}{2021}\natexlab{}.
\newblock \showarticletitle{Learning transferable visual models from natural
  language supervision}. In \bibinfo{booktitle}{\emph{International Conference
  on Machine Learning}}. PMLR, \bibinfo{pages}{8748--8763}.
\newblock


\bibitem[Rui et~al\mbox{.}(2000)]%
        {rui2000automatically}
\bibfield{author}{\bibinfo{person}{Yong Rui}, \bibinfo{person}{Anoop Gupta},
  {and} \bibinfo{person}{Alex Acero}.} \bibinfo{year}{2000}\natexlab{}.
\newblock \showarticletitle{Automatically extracting highlights for TV baseball
  programs}. In \bibinfo{booktitle}{\emph{Proceedings of the eighth ACM
  international conference on Multimedia}}. \bibinfo{pages}{105--115}.
\newblock


\bibitem[San~Pedro et~al\mbox{.}(2009)]%
        {san2009you}
\bibfield{author}{\bibinfo{person}{Jose San~Pedro}, \bibinfo{person}{Vaiva
  Kalnikaite}, {and} \bibinfo{person}{Steve Whittaker}.}
  \bibinfo{year}{2009}\natexlab{}.
\newblock \showarticletitle{You can play that again: exploring social
  redundancy to derive highlight regions in videos}. In
  \bibinfo{booktitle}{\emph{Proceedings of the 14th international conference on
  Intelligent user interfaces}}. \bibinfo{pages}{469--474}.
\newblock


\bibitem[Song et~al\mbox{.}(2016)]%
        {song2016click}
\bibfield{author}{\bibinfo{person}{Yale Song}, \bibinfo{person}{Miriam Redi},
  \bibinfo{person}{Jordi Vallmitjana}, {and} \bibinfo{person}{Alejandro
  Jaimes}.} \bibinfo{year}{2016}\natexlab{}.
\newblock \showarticletitle{To click or not to click: Automatic selection of
  beautiful thumbnails from videos}. In \bibinfo{booktitle}{\emph{Proceedings
  of the 25th ACM international on conference on information and knowledge
  management}}. \bibinfo{pages}{659--668}.
\newblock


\bibitem[Song et~al\mbox{.}(2015)]%
        {song2015tvsum}
\bibfield{author}{\bibinfo{person}{Yale Song}, \bibinfo{person}{Jordi
  Vallmitjana}, \bibinfo{person}{Amanda Stent}, {and}
  \bibinfo{person}{Alejandro Jaimes}.} \bibinfo{year}{2015}\natexlab{}.
\newblock \showarticletitle{Tvsum: Summarizing web videos using titles}. In
  \bibinfo{booktitle}{\emph{Proceedings of the IEEE conference on computer
  vision and pattern recognition}}. \bibinfo{pages}{5179--5187}.
\newblock


\bibitem[Sun et~al\mbox{.}(2014)]%
        {sun2014ranking}
\bibfield{author}{\bibinfo{person}{Min Sun}, \bibinfo{person}{Ali Farhadi},
  {and} \bibinfo{person}{Steve Seitz}.} \bibinfo{year}{2014}\natexlab{}.
\newblock \showarticletitle{Ranking domain-specific highlights by analyzing
  edited videos}. In \bibinfo{booktitle}{\emph{European conference on computer
  vision}}. Springer, \bibinfo{pages}{787--802}.
\newblock


\bibitem[Sun et~al\mbox{.}(2016)]%
        {sun2016videoforest}
\bibfield{author}{\bibinfo{person}{Zhida Sun}, \bibinfo{person}{Mingfei Sun},
  \bibinfo{person}{Nan Cao}, {and} \bibinfo{person}{Xiaojuan Ma}.}
  \bibinfo{year}{2016}\natexlab{}.
\newblock \showarticletitle{VideoForest: interactive visual summarization of
  video streams based on danmu data}. In \bibinfo{booktitle}{\emph{SIGGRAPH
  ASIA 2016 symposium on visualization}}. \bibinfo{pages}{1--8}.
\newblock


\bibitem[Truong et~al\mbox{.}(2021)]%
        {truong2021automatic}
\bibfield{author}{\bibinfo{person}{Anh Truong}, \bibinfo{person}{Peggy Chi},
  \bibinfo{person}{David Salesin}, \bibinfo{person}{Irfan Essa}, {and}
  \bibinfo{person}{Maneesh Agrawala}.} \bibinfo{year}{2021}\natexlab{}.
\newblock \showarticletitle{Automatic generation of two-level hierarchical
  tutorials from instructional makeup videos}. In
  \bibinfo{booktitle}{\emph{Proceedings of the 2021 CHI Conference on Human
  Factors in Computing Systems}}. \bibinfo{pages}{1--16}.
\newblock


\bibitem[Wu et~al\mbox{.}(2011)]%
        {wu2011video}
\bibfield{author}{\bibinfo{person}{Shao-Yu Wu}, \bibinfo{person}{Ruck
  Thawonmas}, {and} \bibinfo{person}{Kuan-Ta Chen}.}
  \bibinfo{year}{2011}\natexlab{}.
\newblock \showarticletitle{Video summarization via crowdsourcing}.
\newblock In \bibinfo{booktitle}{\emph{CHI'11 Extended Abstracts on Human
  Factors in Computing Systems}}. \bibinfo{pages}{1531--1536}.
\newblock


\bibitem[Xiong et~al\mbox{.}(2019)]%
        {xiong2019less}
\bibfield{author}{\bibinfo{person}{Bo Xiong}, \bibinfo{person}{Yannis
  Kalantidis}, \bibinfo{person}{Deepti Ghadiyaram}, {and}
  \bibinfo{person}{Kristen Grauman}.} \bibinfo{year}{2019}\natexlab{}.
\newblock \showarticletitle{Less is more: Learning highlight detection from
  video duration}. In \bibinfo{booktitle}{\emph{Proceedings of the IEEE/CVF
  conference on computer vision and pattern recognition}}.
  \bibinfo{pages}{1258--1267}.
\newblock


\bibitem[Yang et~al\mbox{.}(2022)]%
        {yang2022catchlive}
\bibfield{author}{\bibinfo{person}{Saelyne Yang}, \bibinfo{person}{Jisu Yim},
  \bibinfo{person}{Juho Kim}, {and} \bibinfo{person}{Hijung~Valentina Shin}.}
  \bibinfo{year}{2022}\natexlab{}.
\newblock \showarticletitle{CatchLive: real-time summarization of live streams
  with stream content and interaction data}. In
  \bibinfo{booktitle}{\emph{Proceedings of the 2022 CHI Conference on Human
  Factors in Computing Systems}}. \bibinfo{pages}{1--20}.
\newblock


\bibitem[Yao et~al\mbox{.}(2016)]%
        {yao2016highlight}
\bibfield{author}{\bibinfo{person}{Ting Yao}, \bibinfo{person}{Tao Mei}, {and}
  \bibinfo{person}{Yong Rui}.} \bibinfo{year}{2016}\natexlab{}.
\newblock \showarticletitle{Highlight detection with pairwise deep ranking for
  first-person video summarization}. In \bibinfo{booktitle}{\emph{Proceedings
  of the IEEE conference on computer vision and pattern recognition}}.
  \bibinfo{pages}{982--990}.
\newblock


\bibitem[Yow et~al\mbox{.}(1995)]%
        {yow1995analysis}
\bibfield{author}{\bibinfo{person}{Dennis Yow}, \bibinfo{person}{Boon-Lock
  Yeo}, \bibinfo{person}{Minerva Yeung}, {and} \bibinfo{person}{Bede Liu}.}
  \bibinfo{year}{1995}\natexlab{}.
\newblock \showarticletitle{Analysis and presentation of soccer highlights from
  digital video}. In \bibinfo{booktitle}{\emph{proc. ACCV}},
  Vol.~\bibinfo{volume}{95}. Citeseer, \bibinfo{pages}{11--20}.
\newblock


\bibitem[Zamfirescu-Pereira et~al\mbox{.}(2023)]%
        {zamfirescu2023johnny}
\bibfield{author}{\bibinfo{person}{JD Zamfirescu-Pereira},
  \bibinfo{person}{Richmond~Y Wong}, \bibinfo{person}{Bjoern Hartmann}, {and}
  \bibinfo{person}{Qian Yang}.} \bibinfo{year}{2023}\natexlab{}.
\newblock \showarticletitle{Why Johnny can’t prompt: how non-AI experts try
  (and fail) to design LLM prompts}. In \bibinfo{booktitle}{\emph{Proceedings
  of the 2023 CHI Conference on Human Factors in Computing Systems}}.
  \bibinfo{pages}{1--21}.
\newblock


\end{thebibliography}


\clearpage

\appendix

\onecolumn

\section{Example Highlight Graphs}

\begin{figure*}[!htb]
  \centering
  \includegraphics[width=14cm]{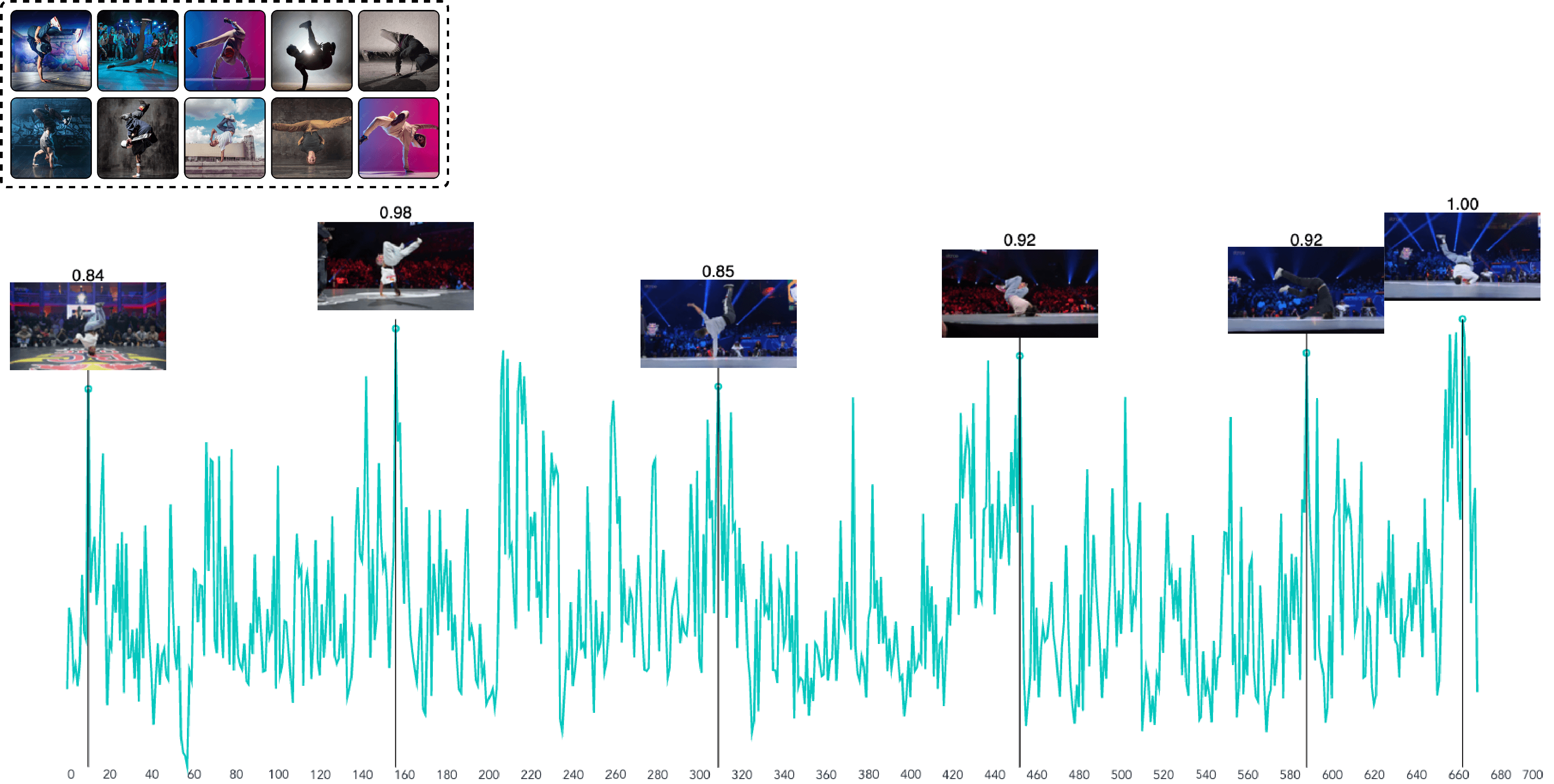}
  \caption{Example highlights from a breakdance competition video. The keyword is \texttt{breakdance}. The photo collection used by Videogenic is shown on the top-left. Videogenic identifies the iconic power moves.}
  \Description{Example highlights from a breakdance competition video. The keyword is breakdance. The photo collection used by Videogenic is shown on the top-left. Videogenic identifies the iconic power moves.}
  \label{fig:example-breakdance}
\end{figure*}

% \begin{figure*}[!htb]
%   \centering
%   \includegraphics[width=14cm]{figures/example-fireworks.png}
%   \caption{Example highlights from a recording of a fireworks show. The keyword is \texttt{fireworks}. The photo collection used by Videogenic is shown on the top-right. Videogenic identifies when the fireworks are in full bloom.}
%   \Description{Example highlights from a recording of a fireworks show. The keyword is fireworks. The photo collection used by Videogenic is shown on the top-right. Videogenic identifies when the fireworks are in full bloom.}
%   \label{fig:example-fireworks}
% \end{figure*}

% \raggedbottom
% \pagebreak

% \begin{figure*}[!htb]
%   \centering
%   \includegraphics[width=14cm]{figures/example-wedding.png}
%   \caption{Example higlights from a wedding video. The keyword is \texttt{wedding}. The photo collection used by Videogenic is shown on the top-left. Videogenic identifies the highlight moments of the couple.}
%   \Description{Example higlights from a wedding video. The keyword is wedding. The photo collection used by Videogenic is shown on the top-left. Videogenic identifies the highlight moments of the couple.}
%   \label{fig:example-wedding}
% \end{figure*}

\begin{figure*}[!htb]
  \centering
  \includegraphics[width=14cm]{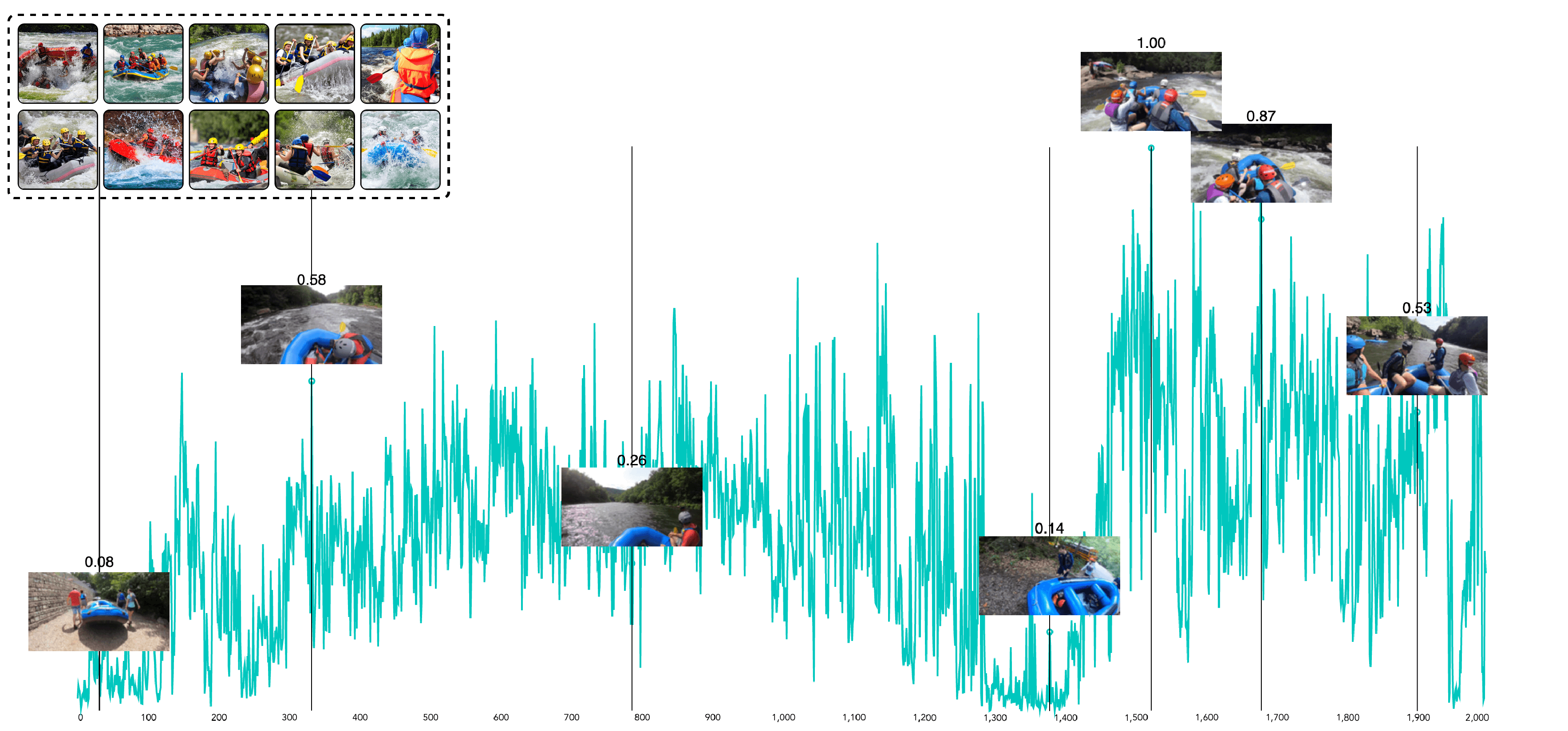}
  \caption{Example video frames and highlight scores within around 30 minutes video footage from a rafting trip. The video clips are recorded by one of the authors using an action camera. The keyword is \texttt{rafting}. The photo collection used by Videogenic is shown on the top-left. We see that Videogenic scores the whitewater moments (i.e., raft going through the river rapids) more highly.}
  \Description{Example video frames and highlight scores within around 30 minutes video footage from a rafting trip. The video clips are recorded by one of the authors using an action camera. The keyword is rafting. The photo collection used by Videogenic is shown on the top-left. We see that Videogenic scores the whitewater moments (i.e., raft going through the river rapids) more highly.}
  \label{fig:example-rafting}
\end{figure*}

\raggedbottom
\pagebreak

\section{\rev{Comparison Against CLIP with Descriptive Text Prompts}}

\rev{In Section \ref{section:human-evaluation}, we compare Videogenic against a CLIP text-image baseline prompted with the topic of the highlight video (e.g., "skateboarding"). In this section, we examine prompting the baseline with a variety of more descriptive prompts, even including prompts that contain domain-specific knowledge (e.g., "kickflip"). Table \ref{tab:highlights-with-videogenic-and-clip-descriptive} shows several prompts and retrieved highlight moments. We show more prompts and retrieved highlight moments in Tables \ref{skateboarding-descriptive-clip} and \ref{wedding-descriptive-clip}. We observe that Videogenic still produces higher quality results than CLIP prompted with descriptive text prompts.}

\rev{We discuss the findings using results shown in Table \ref{skateboarding-descriptive-clip}. First, we test prompts that explicitly describe the task of finding highlights (e.g., "skateboarding highlight" and "photogenic skateboarding"). We also test prompts that implicitly leverage highlight-related cues (e.g., "skateboarding tiktok" and "professional photograph of skateboarding"). We see that these prompts generally fail to retrieve high-quality highlight moments. Second, assuming that the user has domain knowledge of the activity, we test prompts that include domain-specific keywords (e.g., "skateboarding focus on the highlight trick", "skateboarding flip in the air", and "kickflip"). We see that while these prompts help identify impressive skateboarding moments (i.e., the trick), they are often aesthetically unpleasing in terms of composition and framing. Third, we prompt for an aesthetic shot (e.g., "aesthetic skateboarding highlight shot"). We take a step further, assuming that the user has domain knowledge of what shots are most aesthetically pleasing (e.g., "skateboarding close up shot" and "skateboarding low angle shot"). We see that these prompts capture aesthetic but less interesting shots. Finally, we test prompts that include domain knowledge of moments and shot composition (e.g., "close up shot of skateboarding flip in the air" and "low angle shot of a kickflip"). We see that these prompts still produce subpar highlight moments compared to Videogenic.
Overall, we can conclude that prompting for good highlight moments is challenging. This aligns with prior research findings of how people generally struggle with crafting good prompts \cite{zamfirescu2023johnny}. We see that even descriptive text prompts containing rich domain knowledge such as impressive actions and shot specifications are far less capable of identifying highlight moments, compared with leveraging the visual priors encoded within professional photographs.}

\begin{table}[H]
\resizebox{\textwidth}{!}{%
\begin{tabular}{|l|l|l|l|}
\hline
\textbf{Videogenic} & \textbf{"professional photograph of ..."} & \textbf{Descriptive prompt} & \textbf{Domain knowledge prompt} \\ \hline
\textit{<professional photographs>} & "professional photograph of skateboarding" & "skateboarding flip in the air"  & "kickflip" \\ \hline
\includegraphics[width=0.2\columnwidth]{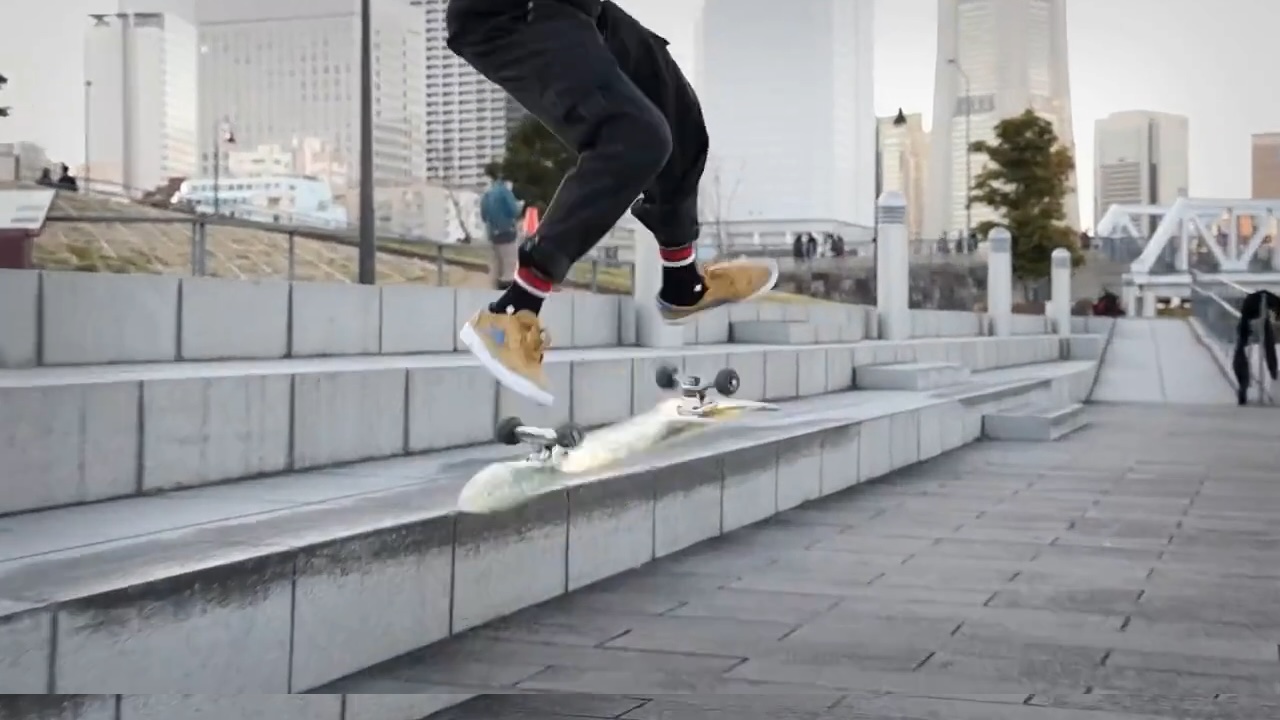} & \includegraphics[width=0.2\columnwidth]{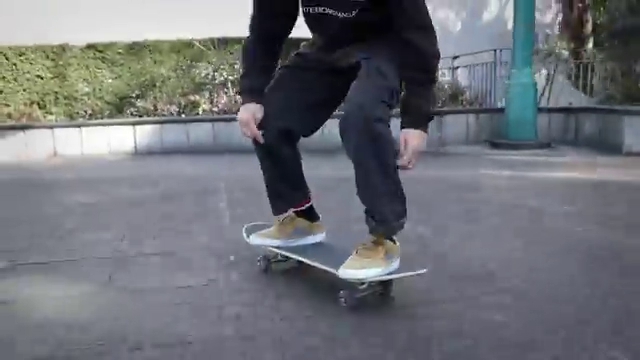} & \includegraphics[width=0.2\columnwidth]{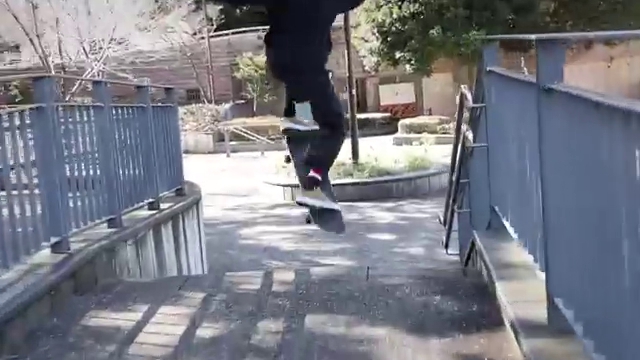} & \includegraphics[width=0.2\columnwidth]{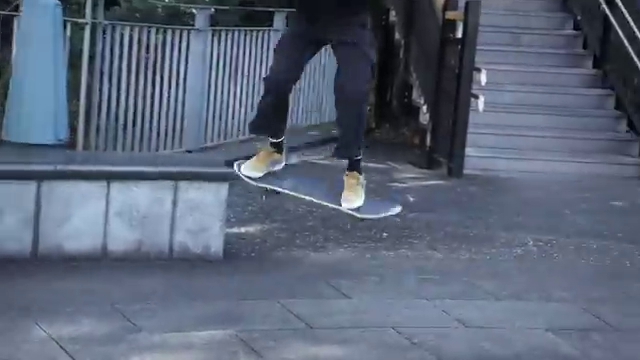}  \\ \hline
\textit{<professional photographs>}           & "professional photograph of a wedding" & "wedding focusing on the couple" & "officiant address" \\ \hline
\includegraphics[width=0.2\columnwidth]{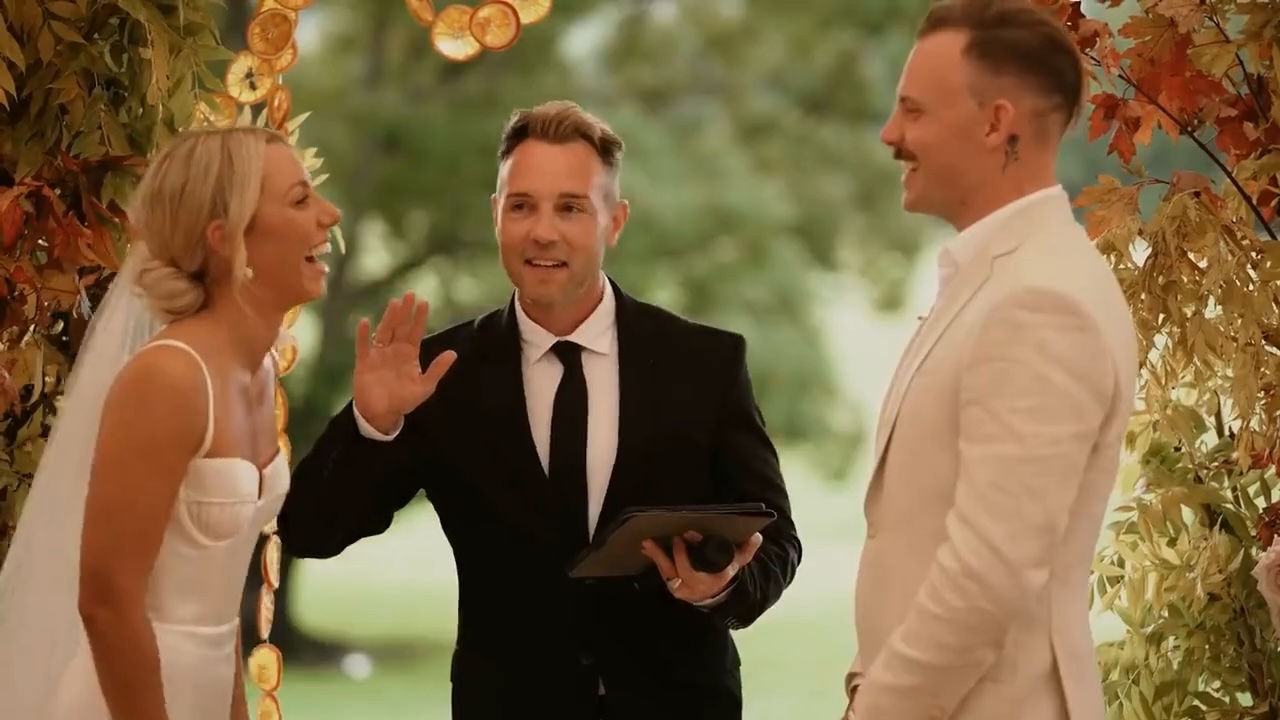}  & \includegraphics[width=0.2\columnwidth]{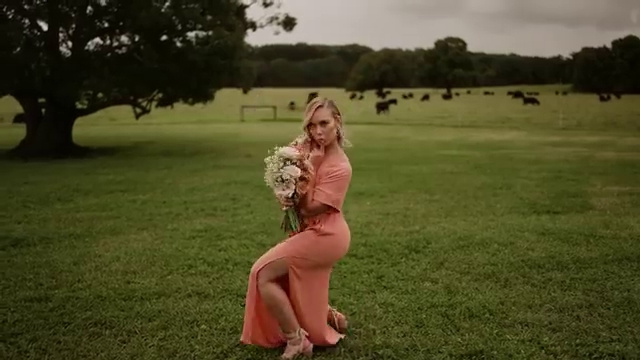} & \includegraphics[width=0.2\columnwidth]{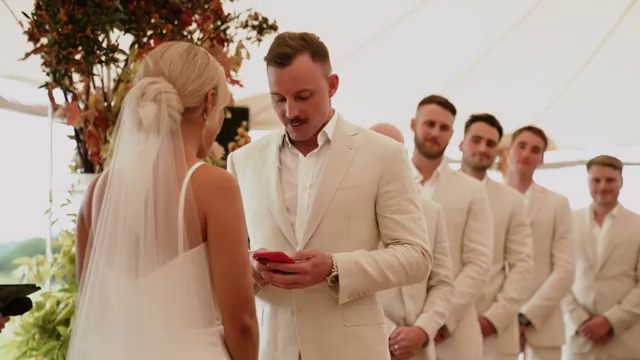}  & \includegraphics[width=0.2\columnwidth]{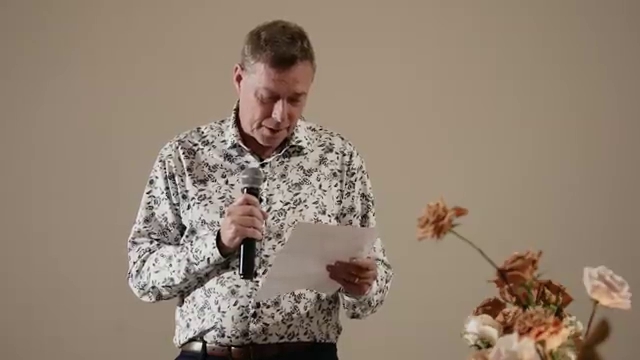} \\ \hline
\end{tabular}%
}
\caption{\rev{Highlight moments identified using Videogenic versus CLIP prompted with a descriptive prompt, a domain knowledge prompt (e.g., the name of a skateboard trick), and "professional photograph of <domain>".}}
\label{tab:highlights-with-videogenic-and-clip-descriptive}
\end{table}

\raggedbottom
\pagebreak

\begin{longtable}{|l|l|}
\hline
\textbf{Query}                                & \textbf{Highlight moment} \\ \hline
\endfirsthead
\endhead
\textbf{Videogenic}                                 & \includegraphics[width=0.28\columnwidth]{figures/videogenic-skateboarding.jpg}                         \\ \hline
skateboarding                                 & \includegraphics[width=0.28\columnwidth]{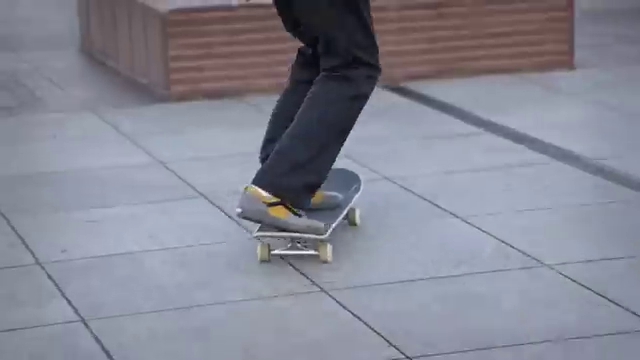}                         \\ \hline
skateboarding highlight                       & \includegraphics[width=0.28\columnwidth]{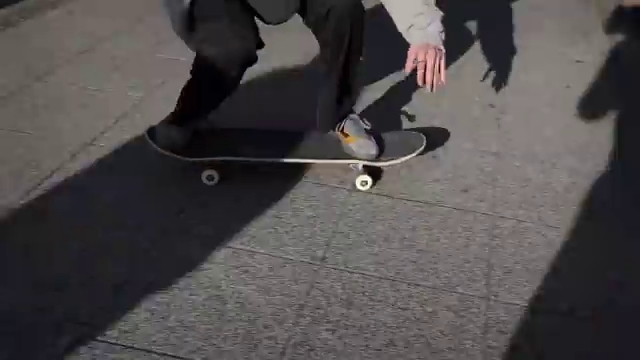}                         \\ \hline
photogenic skateboarding                       & \includegraphics[width=0.28\columnwidth]{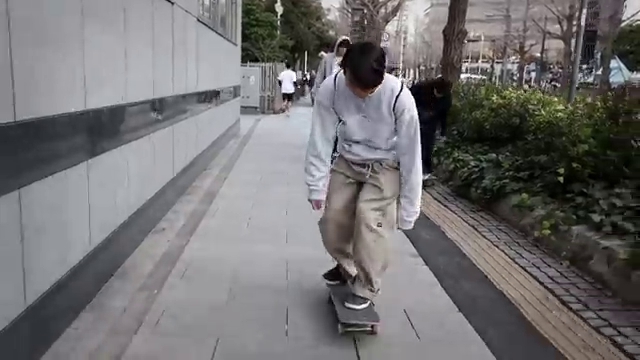}                         \\ \hline
skateboarding tiktok      & \includegraphics[width=0.28\columnwidth]{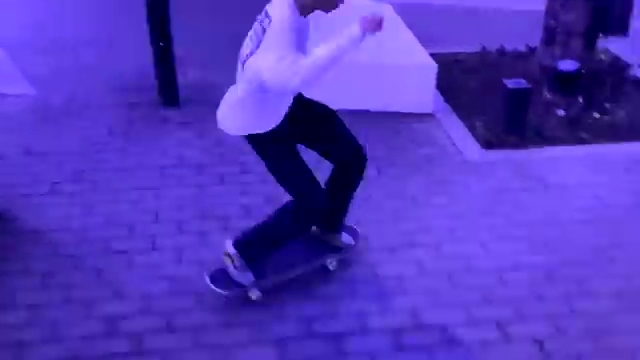}                         \\ \hline
professional photograph of skateboarding      & \includegraphics[width=0.28\columnwidth]{figures/professional-photograph-of-skateboarding.jpeg}                         \\ \hline
skateboarding focusing on the highlight trick & \includegraphics[width=0.28\columnwidth]{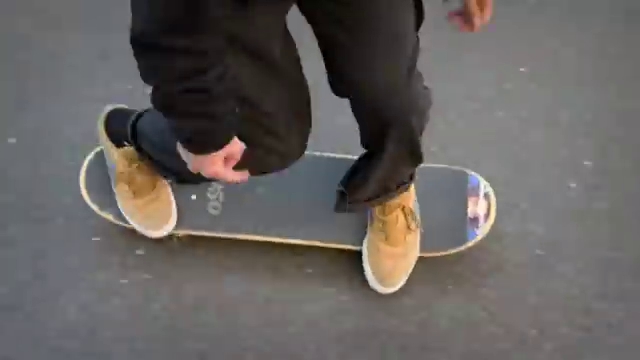}                         \\ \hline
skateboarding flip in the air      & \includegraphics[width=0.28\columnwidth]{figures/skateboarding-flip-in-the-air.jpeg}                         \\ \hline
kickflip                                      & \includegraphics[width=0.28\columnwidth]{figures/kickflip.jpeg}                         \\ \hline
aesthetic skateboarding highlight shot        & \includegraphics[width=0.28\columnwidth]{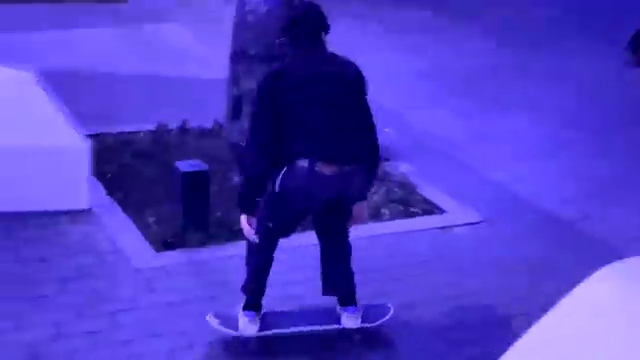}                         \\ \hline
skateboarding close up shot                       & \includegraphics[width=0.28\columnwidth]{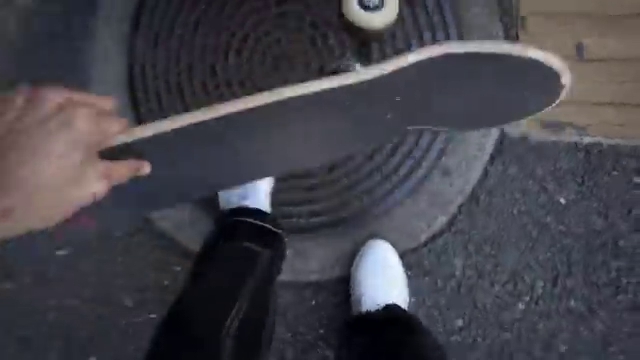}                         \\ \hline
skateboarding low angle shot                  & \includegraphics[width=0.28\columnwidth]{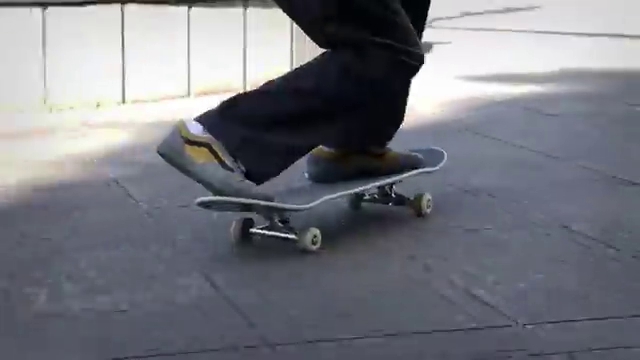}                         \\ \hline
close up shot of skateboarding flip in the air                   & \includegraphics[width=0.28\columnwidth]{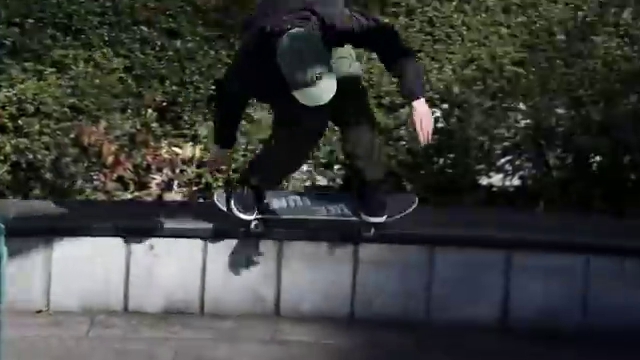}                         \\ \hline
low angle shot of a kickflip                   & \includegraphics[width=0.28\columnwidth]{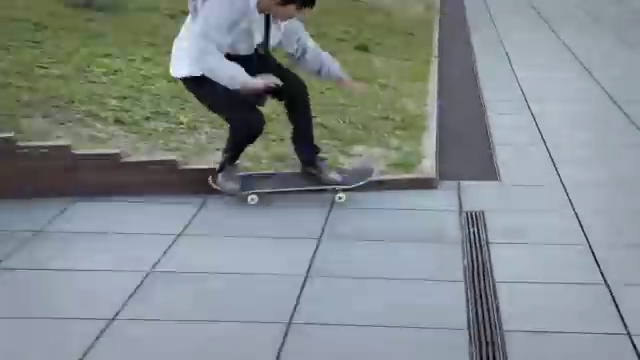}                         \\ \hline
\caption{\rev{Skateboarding highlight moments identified using Videogenic versus CLIP prompted with various text prompts.}}
\label{skateboarding-descriptive-clip}
\end{longtable}

\begin{longtable}{|l|l|}
\hline
\textbf{Query}                                & \textbf{Highlight moment} \\ \hline
\endfirsthead
\endhead
\textbf{Videogenic}                                 & \includegraphics[width=0.28\columnwidth]{figures/videogenic-wedding.jpg}                         \\ \hline
wedding                                 & \includegraphics[width=0.28\columnwidth]{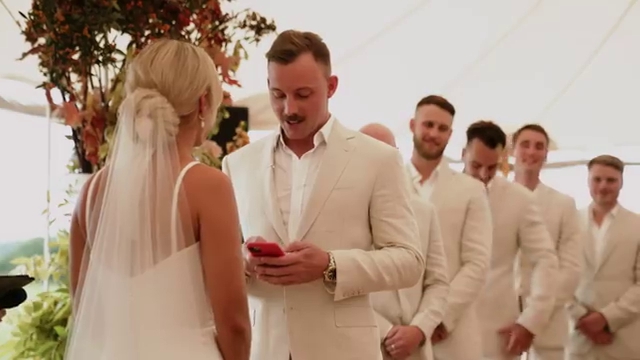}                         \\ \hline
wedding highlight                       & \includegraphics[width=0.28\columnwidth]{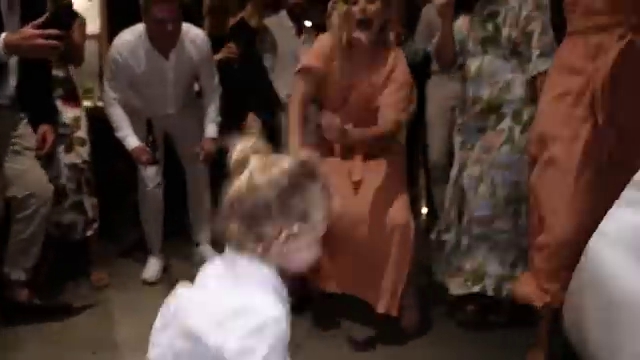}                         \\ \hline
photogenic wedding                       & \includegraphics[width=0.28\columnwidth]{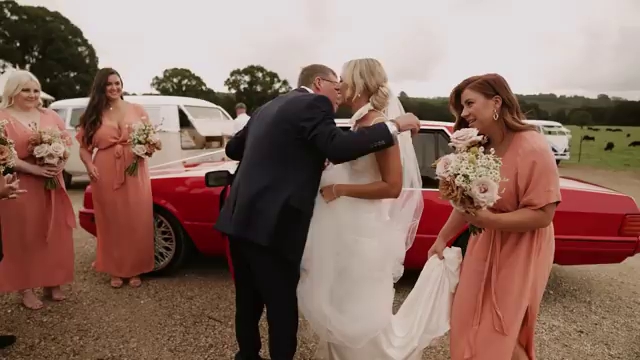}                         \\ \hline
wedding highlight reel      & \includegraphics[width=0.28\columnwidth]{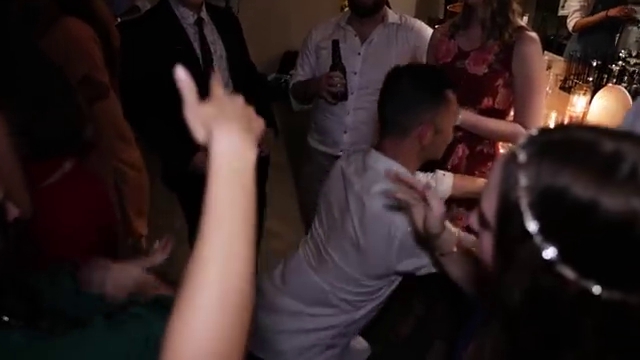}                         \\ \hline
professional photograph of a wedding      & \includegraphics[width=0.28\columnwidth]{figures/professional-photograph-of-a-wedding.jpeg}                         \\ \hline
wedding focusing on the couple & \includegraphics[width=0.28\columnwidth]{figures/wedding-focusing-on-the-couple.jpeg}                         \\ \hline
officiant address                                      & \includegraphics[width=0.28\columnwidth]{figures/officiant-address.jpeg}                         \\ \hline
wedding couple vow      & \includegraphics[width=0.28\columnwidth]{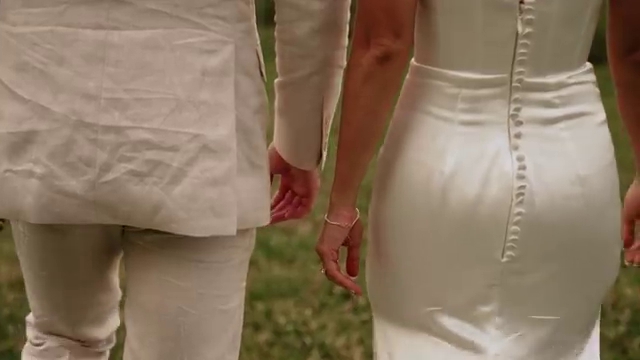}                         \\ \hline
aesthetic wedding shot        & \includegraphics[width=0.28\columnwidth]{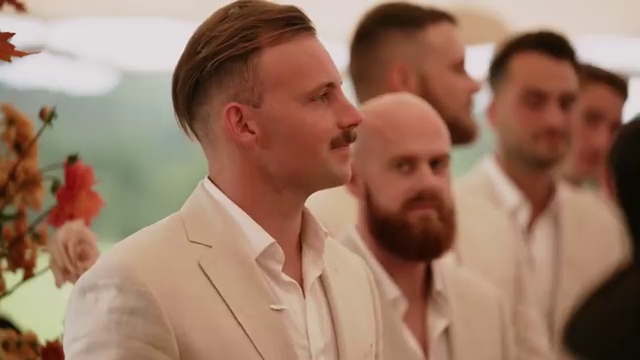}                         \\ \hline
wedding close up shot                       & \includegraphics[width=0.28\columnwidth]{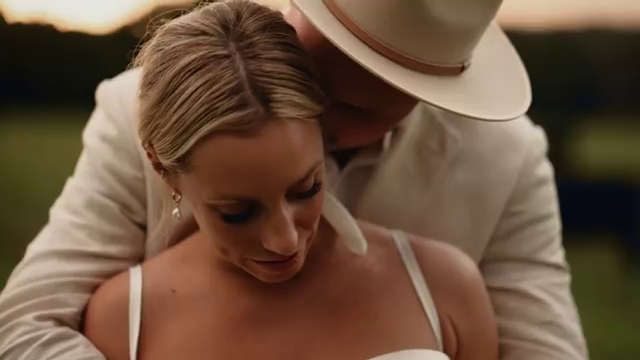}                         \\ \hline
wedding symmetric shot                  & \includegraphics[width=0.28\columnwidth]{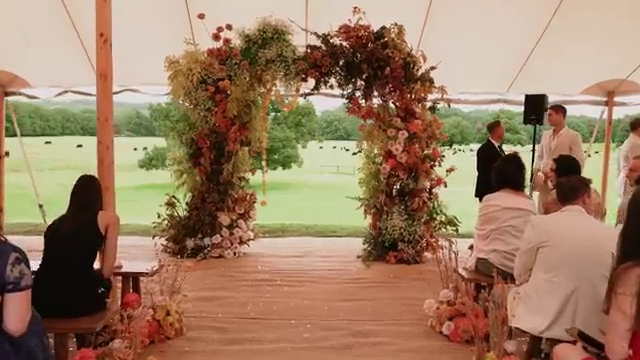}                         \\ \hline
close up shot of couple vow                   & \includegraphics[width=0.28\columnwidth]{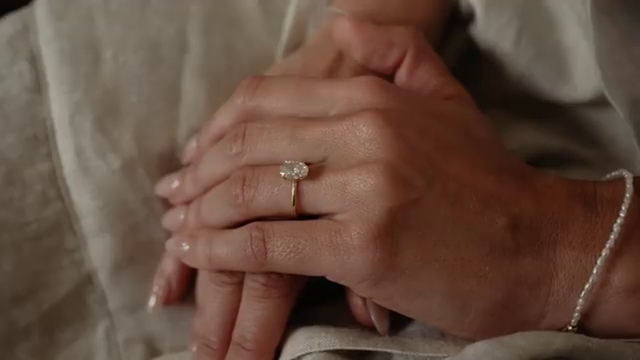}                         \\ \hline
symmetric shot of the couple                   & \includegraphics[width=0.28\columnwidth]{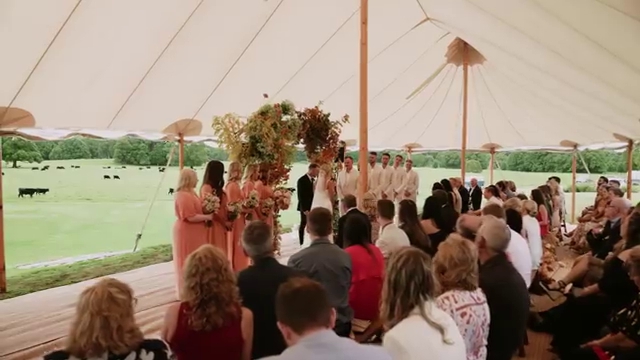}                         \\ \hline
\caption{\rev{Wedding highlight moments identified using Videogenic versus CLIP prompted with various text prompts.}}
\label{wedding-descriptive-clip}
\end{longtable}

\end{document}